%% file: main.tex
\documentclass{article}
\usepackage[preprint]{staix_2026}
\makeatletter\renewcommand{\@noticestring}{}\makeatother

\usepackage[T1]{fontenc}
\usepackage{caption}
\usepackage{subcaption}
\usepackage{tabularx}
\usepackage{microtype}
\usepackage{graphicx}
\usepackage{booktabs,placeins}
\usepackage{float}
\usepackage{amsmath,amssymb}
\usepackage{mathtools}
\usepackage{amsthm}
\usepackage{enumitem}
\usepackage{array}
\usepackage{algorithm}
\usepackage{algorithmic}

\newcommand{\logC}{\operatorname{logC}}
\newcommand{\logB}{\operatorname{logB}}
\newcommand{\logF}{\operatorname{logF}}

\DeclareMathOperator{\logaddexp}{logaddexp}

\newcommand{\mXobs}{\mathcal{X}_{\text{obs}}}

\newcolumntype{L}[1]{>{\raggedright\arraybackslash}p{#1}}
\newcolumntype{R}[1]{>{\raggedleft\arraybackslash}p{#1}}
\newcolumntype{C}[1]{>{\centering\arraybackslash}p{#1}}

\usepackage{hyperref}
\hypersetup{
  pdftitle={Symbolic Density Estimation for Discrete Distributions},
  pdfauthor={Ziwen Liu, Meng Li},
  pdfsubject={},
  pdfkeywords={symbolic density estimation, discrete distributions, probability mass functions}
}

\usepackage[capitalize,noabbrev]{cleveref}
\usepackage[disable,textsize=tiny]{todonotes}

\theoremstyle{plain}

\theoremstyle{definition}

\theoremstyle{remark}

\title{Symbolic Density Estimation for Discrete Distributions}

\author{%
  Ziwen Liu \\
  Rice University \\
  Houston, TX \\
  \texttt{zl166@rice.edu}
  \And
  Meng Li \\
  Rice University \\
  Houston, TX \\
  \texttt{meng@rice.edu}
}

\begin{document}

\maketitle

\begin{abstract}
Discrete probability laws underpin statistical modeling, yet the catalog of interpretable distributions has expanded only gradually through centuries of case-by-case mathematical derivations. We introduce symbolic density estimation (SDE), an unsupervised framework that automatically recovers closed-form probability mass functions by composing elementary analytic operations within a structured search space.
Our method integrates domain-specific structural priors with evolutionary search and a validity-aware inference stage, and it extends to richer distribution families such as zero inflation and finite mixtures. To support systematic evaluation and future research, we contribute a benchmark dataset spanning a broad collection of commonly used discrete distributions. The proposed algorithm recovers all benchmark families with accurate parameter estimates. A real data application shows that it identifies concise and interpretable mixture models that improve goodness-of-fit over standard models.
\end{abstract}

\section{Introduction}
\label{sec:intro}
\input{sections/introduction}

\section{Method}
\label{sec:method}
\input{sections/method}

\section{Experiments}
\label{sec:experiments}
\input{sections/experiments}

\section{Related Work}
\label{sec:related}
\input{sections/related_work}

\section{Conclusion}
\label{sec:conclusion}
\input{sections/conclusion}

\clearpage

\bibliography{references}
\bibliographystyle{plainnat}

\newpage
\appendix
\label{sec:Appendix}
\input{sections/Appendix}

\end{document}

%% file: sections/introduction.tex
Discrete probability distributions are fundamental for modeling count data and categorical outcomes. Traditionally, analysts assume a parametric family and estimate its parameters; however, selecting the appropriate family \textit{a priori} remains a significant challenge and may suffer from model misspecification. Moreover, the dictionary of widely used, analytically tractable distributions remains relatively limited and has expanded mostly through case-by-case constructions~\citep{feller1968introduction}. At the other extreme, flexible nonparametric and neural approaches, such as autoregressive networks and normalizing flows \citep{papamakarios2021, bondtaylor2022}, achieve excellent predictive performance but typically yield \textit{implicit} models rather than closed-form probability mass functions (PMFs). Compact analytic expressions are attractive because they are interpretable, easy to communicate, amenable to mathematical analysis, and often generalize across settings.

We address the problem of automatically recovering a symbolic closed-form PMF that best fits observed discrete data, without prespecifying a family.
Achieving this is non-trivial due to the combinatorially large expression space and the requirement that any proposed formula defines a valid PMF (non-negative and normalized). Unlike standard regression, discrete distributions involve combinatorial constructs like factorials and binomial coefficients.

We propose SDE (symbolic density estimation), a symbolic discovery framework for discrete distributions. The method minimizes a weighted reconstruction error on the log-PMF domain and employs a custom operator set tailored to discrete functional forms. We incorporate automated validity checks for non-negativity and normalization, guiding the search toward interpretable PMFs. The framework is inherently flexible; for instance, by including a simple operator, the system can seamlessly represent mixture distributions and zero-inflated models. Empirical results demonstrate that our approach recovers both standard and non-standard discrete distributions: in synthetic experiments, the system rediscovers correct symbolic PMFs (e.g., Poisson or Negative Binomial) from sample points, while in complex settings with noisy or composite data, it identifies concise expressions outperforming standard families.

Our contributions are: (1) We propose the first symbolic framework for modeling discrete distributions that integrates PMF-validity verification into symbolic discovery, combining weighted log-PMF fitting, built-in normalization checks, and PMF-informed structures to guide symbolic search; (2) we contribute SDEBench, a benchmark dataset spanning a broad collection of commonly used discrete distributions to support systematic evaluation; and (3) we empirically demonstrate that our approach recovers classical, generalized, and structured models, bridging the gap between rigid parametric and black-box methods by automatically recovering interpretable closed-form PMFs from data. 

Code is available at \url{https://github.com/ZiwenLiu2002/SDE-Code}.

%% file: sections/method.tex
\subsection{Problem formulation: symbolic density estimation (SDE)}
\label{sec:Observation Model and Objective}

We consider a univariate discrete target distribution with PMF $p(x)$ on a countable domain $\mathcal{X}$, satisfying $\sum_{x\in\mathcal{X}} p(x)=1$ and $p(x)\ge 0$ for all $x\in\mathcal{X}$. We work in the log domain and model the \emph{log-PMF} $\log p(x)$ directly, which improves numerical stability and turns factorial- and gamma-based factors into additive components that match the canonical structure we aim to discover symbolically.

We seek an analytic expression for $p(x)$ (or equivalently $\log p(x)$) from a hypothesis space $\mathcal{P}$ induced by an operator set $\mathcal{O}$. The set $\mathcal{O}$ combines basic arithmetic operators, elementary functions, and log-domain combinatorial primitives commonly used in discrete probability models \citep{feller1968introduction, johnson1992univariate}. A canonical choice includes arithmetic operators $(+, -, \times, \wedge)$, elementary functions $(\log, \exp, |\cdot|, \sin, \cos)$, and log-domain combinatorial primitives: $\logF(t)=\log\Gamma(t+1)$, $\logC(n,k)=\log\Gamma(n+1)-\log\Gamma(k+1)-\log\Gamma(n-k+1)$, and $\logB(a,b)=\log\Gamma(a)+\log\Gamma(b)-\log\Gamma(a+b)$, where $\Gamma(n+1)=n!$. These primitives express factorials, binomial coefficients, and Beta-function terms common in discrete PMFs while remaining numerically stable for large arguments. 

The symbolic discovery problem in this work induces an extremely large combinatorial search space. Candidate log-PMF expressions are generated by repeatedly composing operators from a fixed operator set under a global expression complexity limit. Even under a modest complexity budget (e.g., $L=20$), the symbolic search space is already enormous; with a vocabulary of roughly a dozen operators, the number of candidate expressions grows exponentially with $L$, yielding upwards of $10^{19}$ possibilities. This growth, driven by rapid increases in admissible expressions with both operator applications and expression depth, constitutes the computational bottleneck in symbolic search.

Enriching $\mathcal{O}$ provides a natural strategy for flexible model structures for $p(x)$ such as zero-inflated and finite mixture distributions; see Section~\ref{sec:extension}. 

\subsection{Estimation of log-PMF with uncertainty quantification}
\label{sub:smoothing}
We estimate the log-PMF from $M$ i.i.d.\ samples and quantify estimation uncertainty to weight the symbolic search targets. Even with exact $\log p(x)$, search remains nontrivial given the vast hypothesis space $\mathcal{P}$ and the requirement that candidate expressions define valid PMFs.

Let $\mXobs = \{x_{\min}, \ldots, x_{\max}\}$ denote the contiguous integer grid spanning the observed range, with $|\mXobs|=K$, and let $c(x)$ be the observed count for each $x \in \mXobs$ (with $c(x)=0$ for integer values in the grid that were not observed). Although the empirical estimator $c(x)/M$ is unbiased where $c(x)>0$, it is not well suited for log-PMF due to zero or nearly zero counts (yielding undefined logarithms) and high variance in low-probability regions. 
To improve numerical stability and mitigate challenges posed by large support, we use the smoothed estimator $\hat p(x) $
\begin{equation}
\hat p(x) = \{c(x) + \alpha\}/\{M + \alpha K\}, \quad 
y(x) = \log \hat p(x), 
\end{equation}
for a small $\alpha > 0$, which corresponds to the posterior mean Bayes estimator under a symmetric Dirichlet prior with concentration parameter $\alpha$ \citep{Gelman2013-gm} and 
additive (Laplace) smoothing \citep{agresti2002categorical}. To further mitigate finite-sample noise in the tails, we optionally apply \emph{support truncation} (hard shrinkage) by restricting the fitting domain 
to bins with sufficient effective counts:
$
\mathcal{X}_{\mathrm{fit}}
= \{\,x \in \mathcal{X} : M\,\hat p(x) \ge \tau\,\},
$
where $\tau$ is a minimum-count threshold. This removes regions that are 
poorly estimated from finite samples. We set $\tau=4$ following standard
heuristics for statistical reliability~\citep{cochran1954}; setting $\tau=0$
disables this truncation.

Empirical log-frequency targets derived from finite samples are inherently
heteroscedastic. By a first-order Delta method~\citep{casella2024statistical},
$\mathrm{Var}[\log \hat p(x)] \approx \{1 - \hat p(x)\}/\{M\,\hat p(x)\}$.
We define the raw weight as the inverse of this variance,
$w_{\mathrm{raw}}(x)=\big(\mathrm{Var}[\log \hat p(x)]\big)^{-1}$. This is the default choice in our experiments; weights from other uncertainty quantification schemes could also be used, such as the posterior variance under the symmetric Dirichlet model above or structured Dirichlet variants \citep{JMLR:v23:21-0335}. 
We normalize the weights by their median:
$w(x) = w_{\mathrm{raw}}(x) / \operatorname{median}_{x \in \mathcal{X}_{\mathrm{fit}}} w_{\mathrm{raw}}(x).$
Together, $y(x)$ and $w(x)$ on $\mathcal{X}_{\mathrm{fit}}$ define a \textit{weighted least-squares} objective that emphasizes reliable observations and down-weights noisy ones.

\subsection{Domain-informed constraints for SDE}
\label{sec:constraints}

To make search tractable, we incorporate two types of domain-informed constraints: (i) \emph{operator complexity profiles} that assign symbolic costs to operators, and (ii) \emph{grammar constraints} that rule out algebraically implausible compositions. Both prune the search space while preserving sufficient expressivity, following a strategy used in physics- and materials-science-informed symbolic regression \citep{Udrescu2020AIFeynman, udrescu2020aifeynman20paretooptimal, Liu2020, Liu2022-kh}.

\paragraph{Operator Complexity Priors.}
To guide symbolic discovery while preserving interpretability, we introduce \emph{operator-level complexity profiles} as soft inductive biases in the symbolic search.
These profiles are used in a multi-start evolutionary strategy and are motivated by recurring structural patterns in discrete log-PMFs.

Many classical discrete distributions admit log-PMF decompositions into two components:
(i) a combinatorial base measure $h(k)$ involving factorial or gamma-function terms (e.g., $\log k!$, $\log\binom{n}{k}$), and
(ii) sufficient statistics $T(k)$ encoding parameter-dependent structure, typically appearing as linear or logarithmic functions of $k$.
In the log-domain, these correspond to additive combinatorial structure and multiplicative exponential-family structure.
This decomposition captures many canonical log-PMF forms and provides a principled basis for structuring symbolic search.

Based on this observation, we define two complementary complexity profiles.
One assigns lower cost to operators associated with combinatorial structure (e.g., $\logC$, $\logB$),
while the other favors exponential-family motifs (e.g., $\logF$, $\log$, multiplication).
Each operator receives a profile-specific symbolic cost, acting as a \emph{soft symbolic prior}~\citep{BRENCE2021107077,schneider2024} during evolutionary selection and mutation.

The profiles guide exploration without restricting expressivity: all operators remain accessible, and global parsimony pressure favors concise expressions. This dual-profile design improves recovery across structurally diverse distributions. Beta-binomial models benefit from profiles permitting $\logB$ terms, while Zipf and Logseries laws favor profiles emphasizing power-law structure. Complete operator-to-complexity assignments are provided in Appendix~\ref{appendix:complexity profile}.

\paragraph{Structural grammar constraints.}

To further control the search space, we restrict the expression grammar to algebraically plausible log-PMF forms: (i) \textbf{Linearity constraint on function arguments.}
    Arguments to special functions such as $\logF$, $\logB$, and $\sin$ are restricted to affine forms.
    This mirrors canonical exponential family structure and prevents deeply nested nonlinear compositions that hinder identifiability. (ii) \textbf{Atomicity constraint on exponents.}
    Exponentiation is restricted to atomic exponents given by a single variable or a constant.
    This discourages compound exponent structures that inflate symbolic complexity and degrade interpretability.

These constraints work together with the complexity control, including the global complexity budget used during search.
Together, they substantially prune the effective search space while preserving expressive power to recover a broad range of discrete distributions.

\subsection{Search and Inference Procedure}
\label{sec:search}
\paragraph{Search algorithm.}
Given the weighted log-frequency targets $y(x)$ on $\mathcal{X}_{\mathrm{fit}}$,
we generate candidate log-PMF expressions via an evolutionary symbolic regression
search under a fixed expression-size budget.
To improve coverage over structurally diverse distributions, we adopt a
multi-start strategy based on the complexity profiles introduced above.
Concretely, we run the same evolutionary search procedure multiple times, each
time using a different profile that assigns distinct per-operator costs.
These costs bias selection and mutation toward different canonical motifs while
keeping all operators accessible, and global parsimony pressure favors concise
expressions under the shared budget.

Within each run, the population is initialized with random expression trees
consistent with the grammar constraints in Section~\ref{sec:constraints}.
Candidates are evolved by mutation and crossover followed
by cost-aware selection: newly proposed expressions are scored by reconstruction
error on $(\mathcal{X}_{\mathrm{fit}}, y, w)$ together with their profile-specific
symbolic cost, and the next generation retains a mix of high-fitness and
low-complexity individuals.
In addition, we periodically apply lightweight algebraic simplification and
re-optimize free numerical constants to improve numerical fit without increasing
symbolic complexity.
All proposals are restricted to satisfy the structural priors.
The search terminates after a fixed number of generations or when the best
candidate stagnates for a preset patience window.

Finally, we pool the candidates produced across profiles and retain a small set
of Pareto-competitive expressions trading off reconstruction error and symbolic complexity \citep{Smits2005}.
These pooled candidates are then passed to the inference stage for validity
checking and parameter refinement. Algorithm~\ref{alg:search} summarizes the full procedure.

\begin{algorithm}[t]
\caption{Evolutionary Search for Log-PMF Expressions}
\label{alg:search}
\vspace{-2pt}
{\scriptsize
\begin{algorithmic}
\STATE {\bfseries Input:} $\mathcal{X}_{\mathrm{fit}},y,w$; $\mathcal{O}$; $\mathcal{G}$; $\{B_j\}_{j=1}^J$;
population $N$, generations $G$, patience $P$, top-$K$
\STATE {\bfseries Output:} candidate set $\mathcal{C}$
\STATE $\mathcal{C}\leftarrow\emptyset$
\FOR{$j=1$ {\bfseries to} $J$}
  \STATE $\mathcal{P}\leftarrow$ random log-PMF expression trees of size $N$ from $(\mathcal{O},\mathcal{G})$ with budget $B_j$
  \STATE $best\leftarrow+\infty,\ stale\leftarrow0$
  \FOR{$t=1$ {\bfseries to} $G$}
    \STATE $\mathcal{Q}\leftarrow$ offspring by mutation/crossover on $\mathcal{P}$; simplify and refit constants
    \STATE $\mathcal{Q}\leftarrow\{e'\in\mathcal{Q}:\mathrm{GrammarOK}(e',\mathcal{G}),\ \mathrm{Cost}(e')\le B_j\}$
    \STATE Evaluate $\ell(e')=\mathrm{WLS}(e';\mathcal{X}_{\mathrm{fit}},y,w)$ for $e'\in\mathcal{Q}$
    \STATE $\mathcal{P}\leftarrow\mathrm{Select}(\mathcal{P}\cup\mathcal{Q})$ using loss--complexity trade-off
    \STATE update $(best,stale)$; \IF{$stale\ge P$} \STATE {\bfseries break} \ENDIF
  \ENDFOR
  \STATE $\mathcal{C}\leftarrow\mathcal{C}\cup \mathrm{TopK}(\mathcal{P},K)$
\ENDFOR
\STATE {\bfseries return} $\mathrm{ParetoFilter}(\mathcal{C})$
\end{algorithmic}}
\vspace{-4pt}
\end{algorithm}

\paragraph{Inference Framework.}
Given the pooled candidate set $\mathcal{C}$ returned by the evolutionary search, we perform a post-search inference stage that filters invalid expressions and refines parameters.

Each candidate is validated in three steps:
\textbf{(i) loss screening}, where expressions with reconstruction error above a fixed threshold are discarded;
\textbf{(ii) probabilistic validity}, requiring approximate normalization $|\sum_{x\in\mathcal{X}_{\mathrm{fit}}} e^{f(x)} - 1| < \epsilon$ and bounded log-mass $\max_{x\in\mathcal{X}_{\mathrm{fit}}} f(x) < \epsilon$;
and \textbf{(iii) complexity control}, which rejects expressions exceeding the profile-specific operator budget.
Candidates passing all checks are pooled and ranked lexicographically, prioritizing lower symbolic complexity and using residual error to break ties, favoring the simplest adequate explanation.

We then apply symbolic canonicalization by rewriting composite operators (e.g., \texttt{logC}, \texttt{logB}, \texttt{logF}) into equivalent $\log \Gamma$ forms using \texttt{sympy}.
The canonicalized expression is decomposed into constant and variable-dependent components to extract structural cues, enabling lightweight structure-based family identification (e.g., a $\log \Gamma(n+1)$ term indicates a Binomial-like form).

For each candidate family, parameters are initialized from identifiable symbolic terms (e.g., $k \log p$) or via moment-based heuristics.
Refinement proceeds via discrete local search: starting from $\theta$, we evaluate multiplicative perturbations $\theta' = \theta \cdot \delta$ over a small grid, selecting updates that minimize the regularized log-PMF RMSE.
This procedure is iterated to convergence. Finally, among all candidates satisfying the loss criterion, we select the expression with minimal symbolic complexity, breaking ties by error.
The resulting symbolic expression and refined parameters constitute the output of our symbolic density estimation pipeline.

\subsection{Extension: finite mixture and zero inflation}

\label{sec:extension}
The framework handles composite model classes by augmenting $\mathcal{O}$ with two primitives, illustrated for finite mixtures and zero-inflated distributions.

\emph{Finite mixtures.} For any base distribution $g(x;\theta)$, the mixture PMF $P(K=k)=\sum_{i=1}^m w_i g(k;\theta_i)$ ($\sum w_i=1$) has log-form $\log\!\sum_i\exp(\log w_i+\log g(k;\theta_i))$. We introduce the binary primitive $\mathrm{logaddexp}(u,v)=\log(e^u+e^v)$, which represents arbitrary mixtures via nested applications; because it is associative, mixtures become binary trees in log-space with no change to the search procedure.

\emph{Zero-inflated models.}
Zero-inflated distributions are widely used for count data with excess zeros. They combine a point mass at zero with a baseline count
distribution $g(x;\theta)$, yielding a composite structure with both atomic and
parametric components. The model is
$
p(x)
= \pi \mathbf{1}_{\{x=0\}}
+ (1-\pi)\, g(x;\theta),
$
which can be written in log-space as
\begin{equation}
\log p(x) = \mathrm{logaddexp}\!\Big(\log \pi + \log \delta_0(x),\; \log(1-\pi) + \log g(x;\theta)\Big).
\end{equation}
To enable this atomic component within a single symbolic expression, we
introduce an additional primitive $\log \delta_0(x)$ encoding a point mass at
zero.

%% file: sections/experiments.tex
We assess the proposed SDE across a diverse suite of classical and complex discrete distributions, along with a real data application with unknown ground truth. 

Our experiments are designed to address the following research questions:

\begin{itemize}[nosep,leftmargin=*]

\item \textbf{RQ1: Recovery and Robustness.} Can SDE consistently recover the correct symbolic form across a wide range of distributions, and what degree of parameter precision does it achieve under finite-sample sampling noise?

\item \textbf{RQ2: Computational Efficiency.} How efficient is the symbolic search process in terms of the number of evolutionary generations and total runtime required to reach the ground-truth formula?

\item \textbf{RQ3: Extensibility to Structural Variations.} Can the framework be extended with minimal adjustments to recover complex distribution variants, specifically zero-inflation and multi-component mixtures?

\end{itemize}

To this end, we first curate a benchmark dataset SDEBench to support systematic evaluation and future research. 

\subsection{SDEBench for evaluating PMF discovery} \label{sec:benchmark}
SDEBench covers 14 base discrete distributions across four categories:
(i) discrete exponential family (Poisson, Binomial, Geometric, Negative-Binomial, Logseries, Hypergeometric, Negative-Hypergeometric);
(ii) power-law and heavy-tailed laws (Zipf, Zipfian, Yule--Simon);
(iii) mixed discrete models (Beta-Binomial, Beta-Negative-Binomial); and
(iv) additional structured distributions (Boltzmann, Discrete-Laplace),
covering classical discrete laws widely used in statistics and machine learning \citep{feller1968introduction,2020SciPy-NMeth} (full catalog in Appendix~\ref{appendix:dist-overview}).
Any base distribution can form an $m$-component mixture or a zero-inflated variant; we test Binomial mixtures with $m\in\{2,3,4\}$ and three zero-inflated models (ZIP, ZINB, ZIG).
We focus on the finite-sample setting (default $M=50{,}000$); we also study small-sample robustness by varying $M\in\{500,1{,}000,5{,}000,10{,}000\}$ across four families and $M\in\{50,\ldots,2{,}000\}$ for the Geometric case (Appendix~\ref{appendix:sample_size_robustness}). Noiseless results are similar and reported in Appendix~\ref{appendix:detailed_results}.

\subsection{Implementation of SDE and metrics}

We implement SDE following Section~\ref{sec:method}. All searches are performed in the log domain, utilizing the canonical operator set $\mathcal{O}$ and the structural constraints described previously. By including the optional primitives $\mathrm{logaddexp}$ and $\log \delta_0(x)$, the framework seamlessly accommodates finite mixtures and zero-inflated models. The core search procedure remains unchanged across these settings; for example, the number of mixture components $m$ is discovered automatically through the resulting symbolic structure rather than being pre-specified. Once a distribution family and its structure are identified, we refine the parameter values using the post-search inference framework detailed in Section~\ref{sec:search}, simultaneously discovering model structure and estimating model parameters. 

\subsection{Results}

Under the finite-sample (noisy) setting, SDE recovers the correct symbolic form (up to algebraic equivalence) for all 14 base distributions in SDEBench. It also recovers the correct nested mixture structure for the three mixture Binomial cases and identifies the correct forms for all three zero-inflated models. This symbolic accuracy is consistent across repeated runs, demonstrating strong resilience to sampling noise. 

We next compare the inferred parameter values to the ground truth. Table~\ref{tab:noisy_main} shows representative results spanning from simple exponential-family forms (Poisson, Neg.\ Binomial, Binomial) to complex combinatorial structures (Beta-Binomial, requiring up to ${\sim}13{,}000$ evolutionary generations in a representative run), and confirms that inferred parameters are typically within a few percent of ground truth.
Complete results for all distributions, organized by structural category, are in Appendix~\ref{appendix:detailed_results}; Appendix~\ref{appendix:param_shift_sub} further confirms recovery under substantially changed parameter settings.

\begin{table}[t]
\centering
\caption{Symbolic recovery under noise ($M=50{,}000$), spanning exponential-family to complex combinatorial distributions.}
\label{tab:noisy_main}
\scriptsize\setlength{\tabcolsep}{4pt}
\begin{tabular}{lcc}
\toprule
\textbf{Dist.} & \textbf{Inferred} & \textbf{True} \\
\midrule
Poisson        & $\hat\lambda=12.01$          & $\lambda=12.0$ \\
Neg.\ Binom.   & $\hat r=9.999,\hat p=0.700$  & $r=10,p=0.70$ \\
Binomial       & $\hat n=10,\hat p=0.300$     & $n=10,p=0.30$ \\
Beta-Bin.      & $\hat n\!=\!100,\hat\alpha\!=\!1.98,\hat\beta\!=\!4.90$ & $n\!=\!100,\alpha\!=\!2,\beta\!=\!5$ \\
\bottomrule
\end{tabular}
\end{table}

\paragraph{Efficiency.}
Figure~\ref{fig:efficiency} shows a bimodal pattern: simple forms (Geometric, Poisson, Zipf) converge in under 400 generations (seconds), while complex combinatorial models (Beta-Binomial, Negative Hypergeometric) require up to $40{,}000$ generations (minutes). Appendix~\ref{appendix:pruning} shows the complexity profiles and grammar constraints reduce the candidate space by over four orders of magnitude.

\paragraph{Small-sample Robustness.}
Table~\ref{tab:sample_size_robustness} evaluates robustness across sample sizes for four structurally diverse families, showing that simpler targets remain recoverable at small sample sizes while more complex ones require more data. Appendix~\ref{appendix:sample_size_robustness} provides detailed recovered expressions and a per-family analysis, including the Geometric case across $M\in\{50,\ldots,2000\}$ (Table~\ref{tab:geom_small_sample}).

\begin{table}[t]
\centering
\caption{Small-sample robustness: log-PMF MSE across sample sizes for four families of increasing structural complexity. Recovered expressions are in Appendix~\ref{appendix:sample_size_robustness}.}
\label{tab:sample_size_robustness}
\scriptsize\setlength{\tabcolsep}{6pt}
\begin{tabular}{lcccc}
\toprule
$M$ & Yule--Simon & Beta-Binomial & ZIP & 3-mix Binomial \\
\midrule
10,000 & 0.0021 & 0.0013 & 0.0097 & 0.0034 \\
 5,000 & 0.0153 & 0.0329 & 0.0271 & 0.0230 \\
 1,000 & 0.0153 & 0.2120 & 0.0301 & 0.1123 \\
   500 & 0.1338 & 0.6720 & 0.1128 & 0.2340 \\
\bottomrule
\end{tabular}
\end{table}

\paragraph{Mixture and Zero-inflated Models (RQ3).}
Tables~\ref{tab:combined_mix} and~\ref{tab:zero-inflated-combined} summarize results for composite models. For Binomial mixtures, component errors remain small at $m=2,3$ and increase moderately at $m=4$; Appendix~\ref{app:mixture_additional} provides full tables up to six components. For zero-inflated models, SDE recovers the unified structure across ZIP, ZINB, and ZIG. SDE extends to these richer structures without altering the core algorithm; Appendix~\ref{appendix:extra_composite} confirms generality on two further families.

\begin{table}[t]
\begin{minipage}[t]{0.44\linewidth}
  \centering
  \caption{Binomial mixture errors ($\times 10^2$).}
  \label{tab:combined_mix}
  \scriptsize\setlength{\tabcolsep}{3pt}
  \begin{tabular}{cll}
  \toprule
  $m$ & $|\Delta p|\times10^2$ & $|\Delta w|\times10^2$ \\
  \midrule
  2 & $(0.41,\ 2.1)$ & $(1.4,\ 1.7)$ \\
  3 & $(0.41,\ 2.1,\ 2.2)$ & $(1.4,\ 1.7,\ 0.26)$ \\
  4 & $(0.13,\ 5.0,\ 9.4,\ 5.9)$ & $(3.7,\ 6.4,\ 17.0,\ 7.3)$ \\
  \bottomrule
  \end{tabular}
\end{minipage}
\hfill
\begin{minipage}[t]{0.52\linewidth}
  \centering
  \caption{Zero-inflated recovery ($M=100{,}000$; larger $M$ used for stable zero-inflation estimation).}
  \label{tab:zero-inflated-combined}
  \scriptsize\setlength{\tabcolsep}{3pt}
  \begin{tabular}{lccc}
  \toprule
  \textbf{Dist.} & \textbf{Params} & \textbf{True} & \textbf{Inferred} \\
  \midrule
  ZIP  & $(\lambda,\pi)$   & $(3.00,0.35)$       & $(2.99,0.35)$ \\
  ZINB & $(r,p,\pi)$       & $(2.20,0.40,0.35)$  & $(2.20,0.40,0.35)$ \\
  ZIG  & $(p,\pi)$         & $(0.30,0.35)$       & $(0.30,0.34)$ \\
  \bottomrule
  \end{tabular}
\end{minipage}
\end{table}

\subsection{Comparative Evaluation with Existing Methods}
\label{sec:baseline-comparison}

We benchmark against five representative baselines (details in Appendix~\ref{Appendix: baseline-details}):
\textbf{MoM} and \textbf{MLE} are classical parametric estimators requiring the correct family \emph{a priori};
\textbf{KDE} \citep{KDE1,KDE2} and \textbf{Pyro} \citep{bingham2019pyro} are nonparametric and neural baselines yielding implicit densities without closed forms;
\textbf{PySR} \citep{PYSR} applies generic symbolic regression without probabilistic constraints.

With the true family given, MoM/MLE serves as oracle benchmark although they cannot discover new symbolic structure. Still, Table~\ref{tab:param_maxerr} shows SDE achieves precision comparable to oracle parametric baselines, without requiring the family to be known.
\begin{table}[t]
\begin{minipage}[t]{0.46\linewidth}
  \centering
  \captionof{figure}{Evolutionary generations to correct symbolic form ($M=50{,}000$; log scale).}
  \label{fig:efficiency}
  \includegraphics[width=\linewidth]{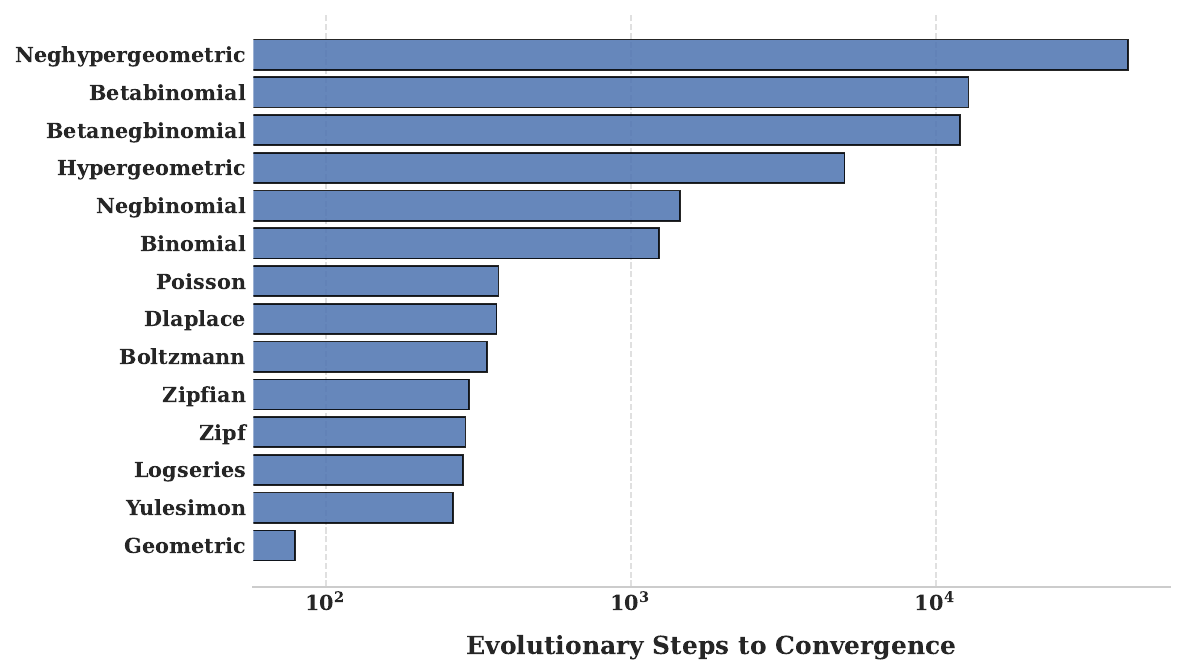}
\end{minipage}
\hfill
\begin{minipage}[t]{0.50\linewidth}
  \centering
  \captionof{table}{MaxErr (\%) for parametric baselines with known families; ``--'' = unsupported.}
  \label{tab:param_maxerr}
  \scriptsize\setlength{\tabcolsep}{4pt}
  \begin{tabular}{lccc}
  \toprule
  Distribution & MoM & MLE & SDE \\
  \midrule
  Poisson        & 0.10 & 0.10 & 0.10 \\
  Binomial       & 0.15 & 0.11 & 0.18 \\
  Geometric      & 0.22 & 0.22 & 0.26 \\
  Neg. Binomial  & 1.75 & 1.70   & 1.76 \\
  Beta-Binomial  & 7.60  & 0.67   & 0.81 \\
  \midrule
  ZIP            & --   & 0.50 & 0.56 \\
  ZIG            & --   & 0.36 & 0.86 \\
  ZINB           & --   & 1.05 & 2.10 \\
  \midrule
  Binomial mixture  & --   & 2.00 & 6.67 \\
  \bottomrule
  \end{tabular}
\end{minipage}
\end{table}

\begin{figure}[t]
\centering
\includegraphics[width=\linewidth]{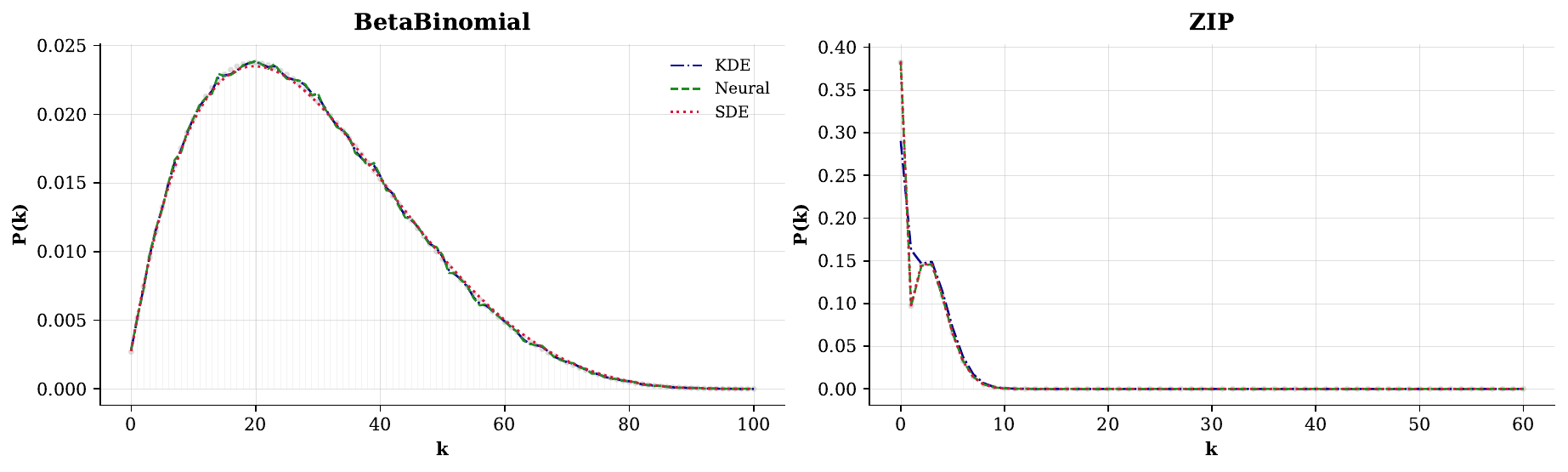}
\caption{PMF estimates for Beta-Binomial (left) and ZIP (right). KDE and Pyro produce non-smooth or boundary-distorted fits; SDE recovers the correct closed-form structure. Full benchmark comparisons are in Appendix~\ref{app:Black-Box Density Estimation Results}.}
\label{fig:kde_pyro_comparison}
\end{figure}

Nonparametric baselines KDE and Pyro lack explicit closed forms and exhibit characteristic failure modes.
Figure~\ref{fig:kde_pyro_comparison} shows two representative cases: KDE produces a boundary artifact near zero for the ZIP distribution, and both KDE and Pyro yield non-smooth, oscillatory fits for the Beta-Binomial. SDE recovers smooth, closed-form PMFs in both cases. A bandwidth-tuning ablation (Appendix~\ref{appendix:kde_ablation}) confirms that tuned KDE still falls well short of SDE.

Tables~\ref{tab:symbolic_comparison} and~\ref{tab:capability_comparison} compare discovery quality and method capabilities. PySR frequently yields structurally invalid expressions; Appendix~\ref{appendix:pysr_controlled} shows this gap persists under a matched budget, and Appendix~\ref{appendix:enumeration} shows brute-force enumeration is computationally infeasible. Among the compared methods, SDE is the only one that requires no predefined family, produces interpretable closed-form expressions, \emph{and} verifies PMF validity through post-search checks (see also Appendix~\ref{Appendix: baseline-details}).

\begin{table}[t]
\centering
\caption{Discovered log-PMF: SDE vs.\ PySR.}
\label{tab:symbolic_comparison}
\scriptsize
\begin{tabularx}{\linewidth}{@{}l X @{}}
\toprule
\textbf{Dist.} & \textbf{log-PMF expression} \\ \midrule
\textbf{Poisson} & \textbf{SDE:} $(x \cdot 2.485) - \text{logF}(x) - 12.01$ \quad
                   \textit{PySR:} $(\sin(\sin(\sin(0.97^x \cdot 6.78))) \cdot -7.39) - 7.65$ \\ \midrule
\textbf{BetaBin.} & \textbf{SDE:} $-22.75 - \text{logB}(101.1 \!- x, 4.02) - \text{logB}(1.01, x \!+ 1.02)$ \quad
                    \textit{PySR:} $(0.85^x + 1.25^{1.02^{x + \cos(x^{2.2})}}) \cdot -2.56$ \\ \bottomrule
\end{tabularx}
\end{table}

\subsection{Real-World Case Study}
\label{sec:real_pbmc}

We evaluate SDE on a human peripheral blood mononuclear cell (PBMC) single-cell RNA-sequencing (scRNA-seq) dataset \citep{Zheng2017-zs}. These data exhibit extreme sparsity ($>$90\% zeros) from biological non-expression or technical dropouts \citep{Pierson2015-uo}, for which ZINB is the standard model \citep{Risso2018-ob}. We estimate the log-PMF of Gene 4046 (82 distinct counts; 2,107 transcripts) and include `MLE+AIC' as a reference that fits a fixed candidate list and selects by the Akaike Information Criterion (AIC). As shown in Tables~\ref{tab:capability_comparison} and~\ref{tab:pbmc_mse_table}, SDE achieves the lowest MSE on this empirical log-PMF while returning an interpretable closed-form PMF.

\begin{table}[t]
\begin{minipage}[c]{0.36\linewidth}
  \centering
  \caption{Capability comparison. ``Family'' = requires known family \emph{a priori}; ``Expr.'' = interpretable output; ``PMF'' = PMF validity verified.}
  \label{tab:capability_comparison}
  \small\setlength{\tabcolsep}{4pt}
  \begin{tabular}{lccc}
  \toprule
  Method & Family & Expr. & PMF \\
  \midrule
  MoM/MLE      & \checkmark & \checkmark & \checkmark \\
  KDE/Pyro     & $\times$   & $\times$   & $\times$   \\
  PySR         & $\times$   & \checkmark & $\times$   \\
  \textbf{SDE} & $\times$   & \checkmark & \checkmark \\
  \bottomrule
  \end{tabular}
\end{minipage}
\hfill
\begin{minipage}[c]{0.60\linewidth}
  \centering
  \caption{PBMC gene 4046 MSE on empirical log-PMF ($\downarrow$ better). ``Interp.''~= symbolic output. MLE+AIC selects from 10 standard families (best: ZINB).}
  \label{tab:pbmc_mse_table}
  \scriptsize\setlength{\tabcolsep}{4pt}
  \begin{tabular}{lccc}
  \toprule
  \textbf{Method} & \textbf{Assumption} & \textbf{Interp.} & \textbf{MSE} \\
  \midrule
  SDE (Ours) & None            & \checkmark & \textbf{0.1263} \\
  PySR       & None            & \checkmark & 0.1406 \\
  Pyro       & Black-box       & --         & 0.2840 \\
  KDE        & Non-parametric  & --         & 0.6196 \\
  MLE+AIC    & 10 fixed families & $\checkmark$ & 0.8509 \\
  MoM        & NB, fixed       & \checkmark & 1.0119 \\
  \bottomrule
  \end{tabular}
\end{minipage}
\end{table}

\begin{figure}[t]
\centering
\includegraphics[width=0.85\linewidth]{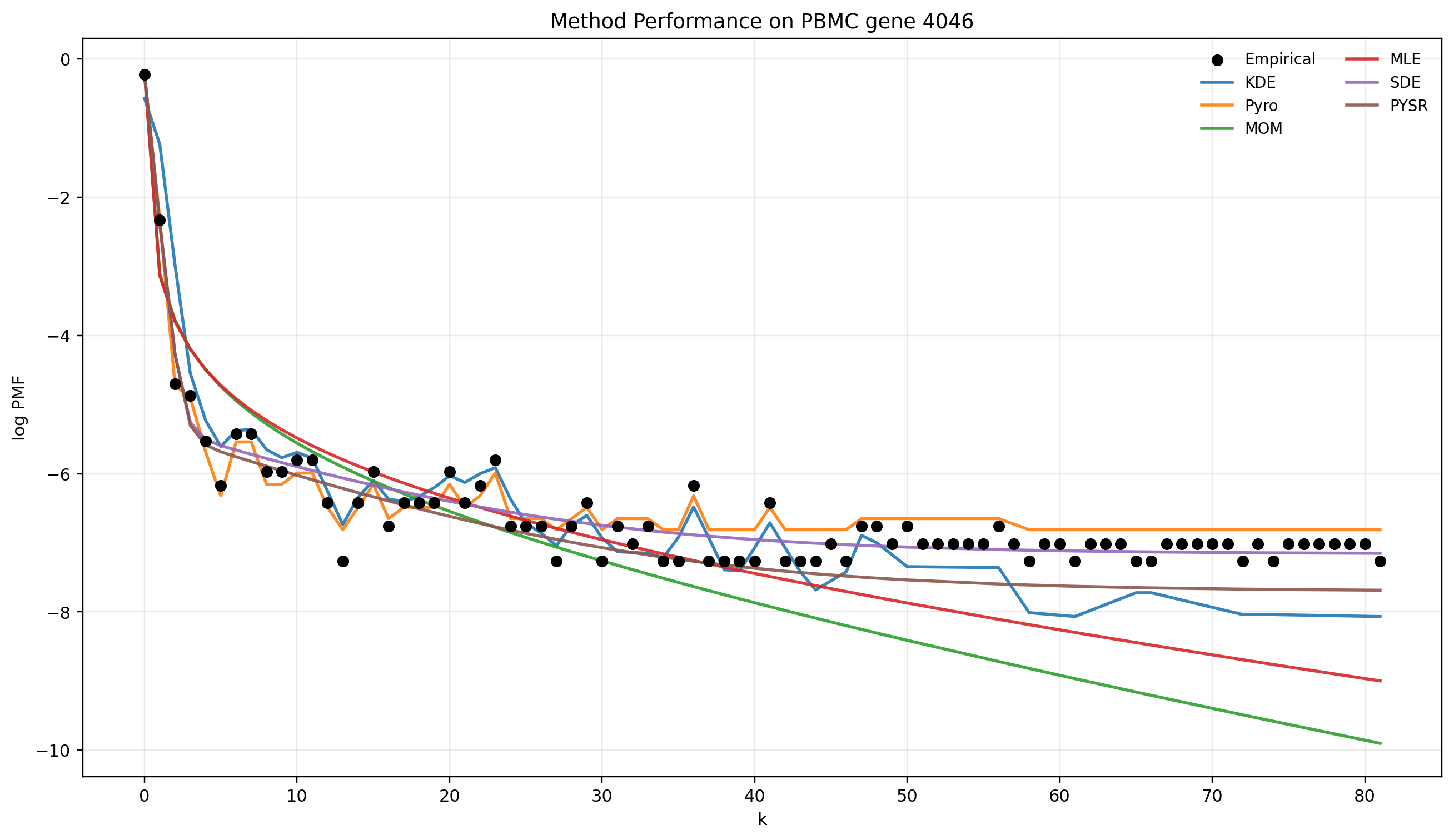}
\caption{Real-data fit on PBMC gene 4046. Black points are the empirical log-PMF over support $k=0,\dots,81$. SDE tracks the data more closely than all baselines while returning a smooth, interpretable closed-form expression.}
\label{fig:pbmc_fit}
\end{figure}

SDE discovers $y(x) = \log\!\bigl(e^{-5.43 - 0.08x} + e^{-0.22 - 2.23x} + e^{-7.16}\bigr)$, a nested \texttt{logaddexp} expression combining two geometric-decay components and a constant baseline term. The fast-decay branch ($\lambda \approx 2.23$) captures transcriptional silence, the slow-decay branch ($\lambda \approx 0.08$) reflects overdispersed expression variability, and the constant term encodes a small residual mass, consistent with biological expectations for scRNA-seq count data \citep{Townes2019-id}. SDE recovers this structure without prior constraints, providing a smooth interpretable formula that surpasses both fixed parametric and black-box models. Figure~\ref{fig:pbmc_fit} shows the fitted log-PMF curves from all methods.

%% file: sections/related_work.tex
\paragraph{Symbolic Regression.}
Symbolic regression (SR) aims to discover closed-form expressions from data and has been approached 
via combinatorial and sparse optimization~\citep{PGE,
Bartlett2020ExhaustiveSR, austel2020, NEUMANN2020123412, McConaghy2011, BRUNTON2016710}, genetic programming~\citep{Koza1992, Poli2008,
Vladislavleva2009, Uy2011, Kronberger2019, PYSR}, Bayesian formulation~\citep{jin2020, bomarito2025}, and model-free variable selection~\citep{Ye2024-og,pan_sr}. Recent advancements also include deep learning, reinforcement learning, and hybrid variants that combine
neural guidance with symbolic search~\citep{allamanis2017, dascoli2023,
valipour2021, shojaee2023, Petersen2019, Petersen2022, kim2021,
Champion_2019, mundhenk2021, crochepierre2022, holt2023dgsr,
Udrescu2020AIFeynman}. However, these \textit{supervised} methods treat the target as a generic function, making it difficult to enforce 
strict PMF-specific structures and distributional constraints like
non-negativity and normalization.

\paragraph{Density Estimation.} Traditional parametric families~\citep{feller1968introduction,casella2024statistical} offer interpretability but suffer from limited capacity and potential model misspecification. In contrast, nonparametric methods and recent neural models, such as autoregressive models and normalizing flows \citep{germain2015,uria2016nade,papamakarios2017maf,dinh2017,papamakarios2021,bondtaylor2022,chen2023,campbell2024dfm} along with their discrete adaptation \citep{tran2019,hoogeboom2019}, 
provide flexibility but often yield black-box representations without interpretable closed-form PMFs. Some recent work has explored symbolic representations for density estimation.
ISR \citep{tohme2024isrinvertiblesymbolicregression} integrates SR into normalizing flows to define densities implicitly via change-of-variable, while MESSY \citep{tohme2024messyestimationmaximumentropybased} derives distributions from moment constraints via maximum entropy, bypassing the discovery of distributional structure directly from data. Overall, existing methods either sacrifice explicit form for flexibility or rely on rigid predefined structures. Consequently, the automatic discovery of interpretable symbolic PMFs from noisy data remains largely unexplored and an open challenge~\citep{papamakarios2021,karlis2023integer,bondtaylor2022}.

\paragraph{Model selection and symbolic model discovery.}
Classical model selection and model averaging provide principled tools for choosing or combining models from a prespecified candidate class, say $\mathcal{M}$, including information criteria, predictive criteria, Bayes factors, and Bayesian model averaging \citep{akaike1973information,akaike1974new,geisser1979predictive,aitkin1991posterior,gelfand1994bayesian,hoeting1999bayesian,claeskens2008model}. This perspective is especially natural in an $\mathcal{M}$-closed setting \citep{bernardo1994bayesian}, where the analyst assumes that the true distribution belongs to a fixed and relatively small collection of plausible candidate families. Related work further studies model comparison and weighting under model misspecification or $\mathcal{M}$-open settings \citep{clyde2013bayesian,li2020comparing}. By contrast, SDE is aimed at symbolic model discovery over a combinatorially large expression space induced by operator compositions. Even under modest complexity constraints, the number of candidate expressions grows extremely rapidly, making the problem fundamentally different from standard fixed-family selection. We therefore view SDE and classical model selection as addressing related but distinct goals: the former emphasizes model discovery in a large symbolic space, whereas the latter emphasizes selection within a prespecified collection of candidate families.

%% file: sections/conclusion.tex
We introduced a symbolic density estimation framework to recover discrete probability mass functions from data. By operating in the log-probability domain with an interpretable set of primitive operators, our approach provides a principled alternative to manual parametric modeling and flexible but opaque black-box density estimators. The framework's core strength lies in its extensibility; complex distributional structures, such as finite mixtures and zero-inflated models, are recovered via minimal operator extensions without modifying the underlying search procedure, loss function, or constraints. These expressions reveal the compositional structure of the underlying distributions, supporting the identification of structured, heterogeneous discrete models.

We contribute the SDEBench dataset for evaluating PMF discovery. Empirically, the method recovers classical discrete laws and compositional variants from finite samples, yielding concise formulas that recover the intended symbolic forms in the benchmark. These results demonstrate that symbolic discovery of discrete PMFs is a practical approach for distribution modeling, especially where interpretability and structural insight are essential.

Several future directions remain. First, the present formulation focuses on univariate discrete distributions. Even in this setting, symbolic density estimation involves a combinatorially large search space together with probabilistic validity requirements such as normalization and non-negativity. Broader discrete models could be handled through principled expansion of the operator set. Multivariate distributions could be approached by allowing multiple variables and incorporating structured factorizations, such as graphical or autoregressive forms. A practical intermediate route is to combine SDE with dimension reduction, for example by applying it to one or several linear combinations of the variables in the spirit of single-index or multiple-index models. Conditional distributions could be modeled by allowing symbolic dependence on covariates, while unordered categorical distributions may require different operator designs because their support does not naturally carry the smoothness or ordering structure used for count distributions. Second, continuous density estimation could be pursued by applying the discrete framework to suitably discretized supports, although a fully continuous version would likely require new validity checks and operator designs to ensure non-negativity and integration to one. Finally, uncertainty quantification (UQ) for both symbolic model selection and parameter estimation is an important open problem. Since the current framework focuses on symbolic discovery, formal UQ for this setting would need to account for selection effects induced by the combinatorial search procedure. Possible directions include variability assessment over symbolic candidates, post-selection inference for parameters, and model averaging in Bayesian or frequentist settings.

%% file: sections/Appendix.tex
\section*{Appendix}

\section{Discrete Distribution Families}
\label{appendix:dist-overview}

Table~\ref{tab:distributions} presents a comprehensive glossary of the discrete probability distribution families examined in this work. Each entry provides the standardized log-probability mass function (Log-PMF) alongside the associated parameters and the specific operator sets required for symbolic reconstruction. The 14 distributions span the major classical univariate discrete families catalogued in standard references \citep{johnson1992univariate, feller1968introduction}, covering all primary families (Poisson, Binomial, Geometric, Negative Binomial, Hypergeometric, Logseries, Negative-Hypergeometric) together with power-law laws (Zipf, Yule--Simon), compound distributions (Beta-Binomial, Beta-Negative-Binomial), and additional structured families. Trivial limiting cases such as Bernoulli (a special case of Binomial with $n=1$) and Discrete Uniform are omitted.

\begin{table}[H]
\caption{\textbf{Discrete probability distributions and their symbolic Log-PMF representations.} This table summarizes the ground truth specifications used for the benchmark evaluation.}
\label{tab:distributions}
\vskip 0.10in
\resizebox{\linewidth}{!}{%
\begin{tabular}{lccc}
\toprule
\textbf{Type} & \textbf{Log-PMF} & \textbf{Parameters} & \textbf{Operators} \\
\midrule
zipf & $-a\log(x) - \log(\zeta(a))$ & $a$ & $\log, *, +, -$ \\
zipfian & $-a\log(x) - \log(H_{a,N})$ & $a, N$ & $\log, *, +, -$ \\
logseries & $x\log(p) - \log(x) - \log(-\log(1-p))$ & $p$ & $\log, *, +, -$ \\
geometric & $\log(p) + (x-1)\log(1-p)$ & $p$ & $\log, *, +, -$ \\
dlaplace & $\log(\tanh(a/2)) - a|x - loc|$ & $a, loc$ & $\text{abs}, +, -$ \\
boltzmann & $-\beta x - \log\left( \frac{e^{-\beta}(1 - e^{-\beta N})}{1 - e^{-\beta}} \right)$ & $\beta, N$ & $*, +, -$ \\
poisson & $x\log(\lambda) - \lambda - \log\Gamma(x+1)$ & $\lambda$ & $\log\Gamma, *, -$ \\
negbinomial & $\log C(x+r-1,x) + r\log(p) + x\log(1-p)$ & $r, p$ & $\log C, \log, *, +, -$ \\
yulesimon & $\log(\rho) + \log B(x, \rho+1)$ & $\rho$ & $\log B, \log, +, -$ \\
betanegbinomial & $\log C(x+r-1,x) + \log B(r+\alpha, x+\beta) - \log B(\alpha, \beta)$ & $r, \alpha, \beta$ & $\log C, \log B, +, -$ \\
binomial & $\log C(n,x) + x\log(p) + (n-x)\log(1-p)$ & $n, p$ & $\log C, \log, *, +, -$ \\
hypergeometric & $\log C(K,x) + \log C(N-K, n-x) - \log C(N,n)$ & $K, N, n$ & $\log C, +, -$ \\
neghypergeometric & $\log C(x+r-1,x) + \log C(N-r-x, K-r) - \log C(N, K)$ & $r, N, K$ & $\log C, +, -$ \\
betabinomial & $\log C(n,x) + \log B(x+\alpha, n-x+\beta) - \log B(\alpha, \beta)$ & $n, \alpha, \beta$ & $\log C, \log B, +, -$ \\
\bottomrule
\end{tabular}%
}
\vskip -0.1in
\end{table}

\section{Complexity Profiles}
\label{appendix:complexity profile}
\label{appendix:pruning}
We define two complementary complexity profiles over the operator set to guide the symbolic regression search. In our evolutionary framework, background knowledge is incorporated via operator-specific complexity penalties that act as soft biases. Specifically, operators are assigned to one of two complexity levels (``low'' or ``high'') based on the expected structural family of the target distribution; these labels are relative and can be adjusted according to domain knowledge. By preferring lower-complexity operators, we bias the search towards simpler expressions, while still allowing all operators through genetic variation and parsimony pressure on expression size. High-complexity operators are therefore used sparingly, which helps regularize the search and prevents overly complex, unnecessarily heavy expressions given a fixed expression-size budget.

We adopt two complementary profiles to encode alternative structural hypotheses, broadening exploration while preserving interpretability. This approach can be particularly helpful for distributions with intricate functional forms (e.g., the Beta-binomial), where recovering the correct analytical form may occasionally require higher-complexity operators. Using both profiles reduces cases where a model achieves low numerical error but fails to capture the true structural representation.

\noindent Profile~1 aligns with combinatorial log-PMF families (e.g., Binomial, Hypergeometric, Beta-binomial), relying mainly on operators like \texttt{logC} and \texttt{logB}. In contrast, Profile~2 targets exponential- and gamma-based distributions (e.g., Poisson, Logseries, Zipf), whose log-PMFs are often expressed through \texttt{log}, \texttt{logF}, and multiplicative forms. The assignments of ``low'' and ``high'' complexity levels for each operator under the two profiles are summarized in Table~\ref{tab:complexity-profiles}.

\begin{table}[!ht] 
\centering 
\caption{Complexity profiles.}
\label{tab:complexity-profiles}
\begin{small}
\begin{sc}
\begin{tabular}{lcc}
\toprule
Operator & Profile 1 & Profile 2 \\
\midrule
\texttt{logC} & low  & high \\
\texttt{logB} & low  & high \\
\texttt{logF} & high & low  \\
\texttt{log}  & high & low  \\
\texttt{*}    & high & low  \\
\texttt{+}    & low  & low  \\
\texttt{-}    & low  & low  \\
\texttt{\string^} & high & high \\
\texttt{exp}  & high & high \\
\texttt{abs}  & high & high \\
\texttt{sin}  & high & high \\
\texttt{cos}  & high & high \\
\bottomrule
\end{tabular}
\end{sc}
\end{small}
\end{table}

\paragraph{Search-space pruning.}
To quantify the effect of the proposed restrictions and priors, we perform an exact syntax-space counting analysis under the same operator grammar used by SDE. We report the counts in a cumulative manner. The unconstrained space counts all admissible expressions under the budget with no additional structural restrictions. We then impose \emph{exponent atomicity}, which restricts exponentiation to simple atomic exponents rather than more complex composite forms. Next, we add the \emph{structural argument restrictions}, which constrain the arguments of special operators and disallow unstable nested compositions among them. Finally, we apply the two \emph{soft complexity profiles}, which do not remove operators from the grammar but instead assign different low/high symbolic costs to bias the practically reachable space. The 1/5 low/high costs in this section are used only for the search-space counting analysis; the main experiments follow the implementation defaults.

Under a uniform complexity metric with budget 20, exponent atomicity alone provides only mild pruning, reducing the space from $6.19 \times 10^{19}$ to $3.50 \times 10^{19}$, i.e., a $1.77\times$ reduction. In contrast, adding the structural argument restrictions reduces the space further to $1.01 \times 10^{15}$, corresponding to an approximately $6.14 \times 10^{4}$-fold reduction relative to the unconstrained grammar.

On top of these hard restrictions, the soft complexity priors further shrink the practically reachable space. Using the same low/high operator costs as in the two complexity profiles, with costs 1 and 5 respectively, Profile~1 reduces the restricted space from $1.01 \times 10^{15}$ to $7.06 \times 10^{10}$, an additional reduction of approximately $1.43 \times 10^{4}$. Profile~2 reduces the same restricted space to $1.05 \times 10^{13}$, corresponding to an additional $96\times$ reduction. Overall, the main pruning effect comes from the structural restrictions, while the soft complexity profiles provide a further multi-order-of-magnitude reduction in the search space explored in practice.

\section{Implementation Details for SDE and Baselines}
\label{Appendix: baseline-details}

This section provides implementation details, hyperparameter settings, and computational environments for SDE and the baseline methods evaluated in Section~\ref{sec:baseline-comparison}.

\paragraph{Implementation details for SDE.}
We summarize the fixed implementation choices used for SDE throughout the experiments.

\begin{itemize}[leftmargin=*,nosep]
    \item \textbf{Target construction and weighting.} In the noisy setting, we construct empirical log-frequency targets from multinomial samples, apply Laplace smoothing with $\alpha=0.5$, and restrict fitting to bins whose expected counts exceed a threshold $\tau=4$. To account for heteroscedastic noise, we use the inverse Delta-method approximation to $\mathrm{Var}[\log \hat{p}(x)]$ as the raw weight and normalize these weights by their median over the retained support. Optionally, a damping exponent $\gamma$~\citep{Carroll2017} can be applied via $w(x)\leftarrow w(x)^{\gamma}$ to further compress the weight range; we set $\gamma=1$ throughout, i.e., simple median normalization.

    \item \textbf{Evolutionary search routine.} Symbolic search follows a population-based evolutionary procedure. Candidate expressions are selected according to their reconstruction quality and symbolic complexity, and are updated through stochastic mutation and crossover operations. Mutations perturb numerical constants, replace operators, or alter subtrees, while crossover exchanges subtrees between parent expressions to maintain structural diversity. After these structural updates, free numerical constants are further refined through an explicit optimization loop.

    \item \textbf{Search configuration.} All tasks use the same fixed search setup: 30 parallel populations, 60 candidate expressions per population, 3,000 search iterations, a maximum expression size of 15, explicit numerical refinement of constants with 8 optimization restarts, and Pareto-based candidate generation. These settings are kept fixed across tasks rather than tuned separately for each distribution.

    \item \textbf{Operator library and structural restrictions.} The operator library is fixed across tasks. We use unary operators $\logF$, $\log$, $\exp$, \texttt{abs}, $\sin$, and $\cos$, and binary operators $+$, $-$, $*$, power, $\logB$, and $\logC$. To avoid unstable or uninterpretable symbolic forms, we impose fixed structural restrictions: special operators are not allowed to nest inside one another, nor to take $*$ or power as direct inner forms, and exponentiation is restricted to simple exponents.

    \item \textbf{Complexity profiles.} We use two fixed complexity profiles with user-configurable low/high operator costs. Profile 1 assigns low cost to $\logC$, $\logB$, $+$, and $-$, favoring combinatorial structure. Profile 2 assigns low cost to $\logF$, $\log$, $*$, $+$, and $-$, favoring factorial- and exponential-type structure. These profiles act as soft structural priors and do not change the operator library itself.
    
    \item \textbf{Composite models.} For finite mixtures and zero-inflated models, $\logaddexp$ and $\log\delta_0$ are added as optional primitives, each assigned complexity cost 1.

    \item \textbf{Post-search filtering and selection.} After running the two profiles independently, all candidates that pass the post-search filters are pooled across profiles. We apply four fixed criteria: reconstruction loss below $10^{-3}$; approximate normalization satisfying $\left|\sum_{x \in \mathcal{X}_{\mathrm{fit}}} e^{f(x)} - 1\right| < 10^{-3}$; bounded log-mass with $\max_{x \in \mathcal{X}_{\mathrm{fit}}} f(x) < 10^{-3}$; and operator-count limits of at most three occurrences each for $\logB$, $\logC$, $\logF$, $\log$, $\exp$, \texttt{abs}, $\sin$, $\cos$, and $*$. The final model is selected by prioritizing lower symbolic complexity and breaking ties by fit quality.

    \item \textbf{Parameter inference.} After model selection, parameters are initialized from identifiable symbolic terms and refined in the post-search inference stage. For mixture models, the number of components is inferred from the discovered $\logaddexp$ structure rather than pre-specified.
\end{itemize}

These defaults are used throughout as fixed stabilizing choices rather than tuned separately for each distribution. In particular, $\alpha=0.5$ stabilizes near-empty bins, $\tau=4$ removes clearly unreliable low-count bins, and $\gamma=1$ uses the median-normalized inverse-variance weights without additional damping; a larger $\gamma$ could be beneficial in settings with more extreme heteroscedasticity.

\paragraph{Parametric Estimators: MoM and MLE.}
Classical parametric estimation was performed using distribution-specific estimators under the assumption that the true model family is known \emph{a priori}.

\begin{itemize}[leftmargin=*,nosep]
    \item \textbf{Method of Moments (MoM):} Parameters for Poisson, Binomial, and Geometric distributions were derived from empirical mean and variance. For over-dispersed models (Negative Binomial and Beta-Binomial), we solved for $(\hat{r}, \hat{p})$ and $(\hat{\alpha}, \hat{\beta})$ using the first two empirical moments.
    \item \textbf{Maximum Likelihood Estimation (MLE):} For simple families (Poisson, Binomial, Geometric), MLE coincides with MoM. For Negative Binomial and Beta-Binomial, MLE is obtained by minimizing the negative log-likelihood numerically via L-BFGS-B with parameter bounds, initialized from MoM estimates. For zero-inflated (ZIP, ZIG, ZINB) and mixture models, MLE is computed via EM with an iteration budget of $10^6$; base distribution parameters are updated using weighted empirical means.
\end{itemize}

\paragraph{Black-Box Density Estimators: KDE and Pyro.}
Non-parametric and neural baselines evaluate the trade-off between local fitting flexibility and structural interpretability.

\begin{itemize}[leftmargin=*,nosep]
    \item \textbf{Kernel Density Estimation (KDE):} A Gaussian kernel was implemented using \texttt{scikit-learn} with a bandwidth set to $h = 0.6$ to balance noise suppression with structural fidelity. Estimated densities were normalized over the discrete support using a stable log-softmax.

    \paragraph{Per-dataset KDE bandwidth tuning.}
    \label{appendix:kde_ablation}
    We evaluate whether per-distribution bandwidth tuning closes the gap to SDE. For each distribution, noisy counts are split 80/20 into training and validation; the bandwidth is selected from the grid $\{0.05, 0.075, 0.10, 0.15, 0.20, 0.30, 0.40, 0.50, 0.60, 0.80, 1.00, 1.50, 2.00, 3.00\}$ by minimizing validation error, and KDE is refit on the full counts with the selected bandwidth. Table~\ref{tab:kde_tuned} shows that tuning improves KDE on some distributions, but SDE remains substantially better overall and additionally provides an interpretable closed-form PMF.
    
    \begin{table}[t]
    \centering
    \scriptsize
    \setlength{\tabcolsep}{3pt}
    \caption{Ablation on KDE bandwidth selection under the same setting as in the main paper, comparing fixed-bandwidth and tuned-bandwidth KDE together with Pyro and SDE. The tuned KDE selects the bandwidth separately for each distribution, while all other settings remain unchanged.}
    \label{tab:kde_tuned}
    \begin{tabular}{lcccccc}
    \toprule
    Distribution & KDE Fixed BW & KDE Fixed MSE & KDE Tuned BW & KDE Tuned MSE & Pyro MSE & SDE MSE \\
    \midrule
    Poisson        & 0.60 & 27.81  & 1.50 & 24.72  & 0.65         & 2.20e-03 \\
    Neg. Binomial  & 0.60 & 137.21 & 0.30 & 111.60 & 0.36         & 1.69     \\
    Geometric      & 0.60 & 0.38   & 0.60 & 0.38   & 2.56e-02     & 0.10     \\
    Binomial       & 0.60 & 0.05   & 0.05 & 0.02   & 0.35         & 6.26e-03 \\
    Beta-Binomial  & 0.60 & 0.21   & 0.20 & 0.06   & 3.44e-02     & 2.00e-03 \\
    ZIP            & 0.60 & 721.63 & 2.00 & 459.49 & 1.08         & 1.04e-02 \\
    ZIG            & 0.60 & 211.24 & 0.10 & 171.36 & 0.93         & 0.63     \\
    ZINB           & 0.60 & 711.34 & 1.00 & 703.57 & 0.66         & 0.86     \\
    Binomial mixture  & 0.60 & 0.02   & 0.80 & 0.02   & 1.87e-03     & 6.56e-04 \\
    \bottomrule
    \end{tabular}
    \vskip -0.05in
    \end{table}
    \item \textbf{Neural Categorical Model (Pyro):} A black-box density estimator implemented using the \texttt{Pyro} probabilistic programming framework. The model parameterizes a categorical distribution via a learnable logit vector of size $K$ (corresponding to the support size). Training was conducted using Stochastic Variational Inference (SVI) with the Adam optimizer (learning rate $= 0.1$). The model was optimized over $1,200$ epochs to maximize the multinomial log-likelihood of the empirical counts, effectively serving as a high-capacity neural baseline for discrete density estimation.
\end{itemize}

\paragraph{Generic Symbolic Regression: PySR.}
We compare against PySR~\citep{PYSR} (v0.16.0), a generic symbolic regression baseline, with a search budget of $3,000$ iterations, $30$ parallel populations, $60$ candidate expressions per population, and maximum expression complexity of $15$ nodes. The operator set is $\{+, -, *, \wedge, \log, \exp, \text{abs}, \sin, \cos\}$. Unlike SDE, PySR performs an unconstrained search without enforced non-negativity, normalization, or specialized combinatorial primitives, but receives the same support locations ($X_{\text{fit}}$) and empirical log-PMF targets ($y$).

\paragraph{Computational Environment.}
All experiments were executed on a workstation with the following hardware and software configuration:
\begin{itemize}[leftmargin=*,nosep]
    \item \textbf{Hardware:} Dual \textbf{NVIDIA GeForce RTX 2080 Ti} GPUs (11GB VRAM each).
    \item \textbf{Driver \& CUDA:} NVIDIA Driver version \textbf{550.54.15} with \textbf{CUDA 12.4}.
    \item \textbf{Software Stack:} Python 3.9, PyTorch 2.1 (for the Pyro baseline), and Julia 1.10.2 (backend for PySR).
\end{itemize}

\paragraph{Evaluation protocol.}
For the benchmark experiments, the noisy target is constructed by drawing finite samples from the underlying PMF and applying additive smoothing to obtain an empirical log-PMF target. SDE is fitted on one such realization, and fit quality is measured on an independently reconstructed noisy target generated from a separate sample of the same underlying distribution using the same sampling-and-smoothing procedure.

For the PBMC case study, the reported MSE is computed with respect to the empirical log-PMF estimated from the filtered gene-count distribution used in the analysis. This example is therefore intended to illustrate fit quality and interpretability of the discovered expression on the observed empirical distribution.

\paragraph{Choice of WLS over MLE.}
The evolutionary search targets the log-PMF function directly, fitting it as a symbolic expression. WLS on the log-PMF is natural in this setting: it operates on the same domain as the search candidates and avoids evaluating the normalization constant $\sum_x e^{f(x)}$ at every candidate expression during search. Maximum likelihood would require this sum at each evaluation step, which is computationally prohibitive across the large candidate set. Validity (approximate normalization) is instead enforced in the post-search inference stage.

\paragraph{PMF validity and normalization.}
PMF validity is handled in a post-search inference stage. During evolutionary search, candidate expressions are ranked by reconstruction error and symbolic complexity. The pooled candidates are then filtered using probabilistic validity checks, including approximate normalization and bounded log-mass. This design preserves useful intermediate expressions during search while ensuring that the final reported model satisfies the required PMF constraints.

\section{Compositional Properties of the \texttt{logaddexp} Operator}
\label{appendix:mixture}

\subsection{From Binary to Multi-Branch Log-Sum-Exp}

We begin by recalling the definition of the binary log-sum-exp operator,
\begin{equation}
\mathrm{logaddexp}(a,b) := \log\left(e^{a} + e^{b}\right).
\end{equation}
Although this operator is binary, it can be extended to represent log-sums over an arbitrary number of terms through iterative composition.

Given scalars $a_1,\dots,a_n$, define
\[
\begin{aligned}
L_2 &:= \mathrm{logaddexp}\big(a_1,a_2\big),\\
L_k &:= \mathrm{logaddexp}\big(L_{k-1},a_k\big), \qquad k=3,\dots,n.
\end{aligned}
\]

By straightforward induction, this construction satisfies
\begin{equation}
L_n = \log\left(\sum_{i=1}^{n} e^{a_i}\right).
\end{equation}
We therefore define the $n$-ary log-sum-exp operator implicitly as
\begin{equation}
\mathrm{multilogaddexp}(a_1,\dots,a_n) := L_n.
\end{equation}

This observation establishes that iterated compositions of the binary \texttt{logaddexp} operator are expressively equivalent to an explicit $n$-ary log-sum-exp, and hence sufficient to represent finite mixtures with an arbitrary number of components.

\subsection{Closure Under Addition of \texttt{logaddexp} Expressions}

We next show that expressions involving \texttt{logaddexp} are closed under addition, a property that is essential for symbolic composition.

Let
\[
U(k)=a_1k+b_1,\;
V(k)=a_2k+b_2,\;
W(k)=a_3k+b_3,\;
Z(k)=a_4k+b_4.
\]

and consider the sum
\begin{equation}
S(k) = \mathrm{logaddexp}(U,V) + \mathrm{logaddexp}(W,Z).
\end{equation}
Expanding each term yields
\begin{equation}
S(k) = \log\!\left[(e^{U}+e^{V})(e^{W}+e^{Z})\right]
= \log\!\left(e^{U+W}+e^{U+Z}+e^{V+W}+e^{V+Z}\right).
\end{equation}
This expression can again be written in terms of nested \texttt{logaddexp} operators:
\begin{equation}
S(k)
= \mathrm{logaddexp}\Big(
\mathrm{logaddexp}(U+W, U+Z),
\mathrm{logaddexp}(V+W, V+Z)
\Big).
\end{equation}

Thus, the sum of two \texttt{logaddexp} expressions can always be rewritten as a single \texttt{logaddexp} tree, demonstrating closure of this representation under addition.

\subsection{Preservation of Affine Structure Under Composition}

In many of the distributions considered in this work, the arguments to \texttt{logaddexp} are affine functions of the variable $x$.
We therefore analyze how affine structure behaves under composition.

Assume each branch takes the form
\begin{equation}
a_i(x) = \alpha_i x + \beta_i,
\qquad
k_j(x) = \kappa_j x + \lambda_j.
\end{equation}
Then any expanded branch produced by composition satisfies
\begin{equation}
a_i(x) + k_j(x)
= (\alpha_i + \kappa_j) x + (\beta_i + \lambda_j),
\end{equation}
which remains affine in $x$.

Consequently, iterative application of \texttt{logaddexp} preserves the affine-in-$x$ structure of all branches, with parameters combining additively.
This property ensures that the resulting symbolic expressions remain interpretable, with each branch corresponding to a distinct mixture component in the log-domain.

\subsection{Growth of Non-Constant Branches}

Finally, we briefly characterize how the number of non-constant branches evolves under composition.
Let
\[
S(x) = \mathrm{multilogaddexp}\!\big(a_1(x),\dots,a_n(x)\big),
\qquad
K(x) = \mathrm{logaddexp}\!\big(k_1(x),k_2(x)\big).
\]

and assume that exactly $m$ of the functions $a_i(x)$ are non-constant (i.e., $\alpha_i \neq 0$).

When both $k_1$ and $k_2$ are non-constant, composition produces $2n$ non-constant branches.
If exactly one of $k_1$ or $k_2$ is constant, the number of non-constant branches becomes $n+m$.
In both cases, branch growth is controlled and remains linear in the number of existing components.

This observation explains why symbolic expressions involving \texttt{logaddexp} remain manageable even as mixture complexity increases.

The results in this appendix establish several structural properties of the \texttt{logaddexp} operator that are directly relevant to symbolic density estimation.
Although \texttt{logaddexp} is a binary primitive, it is sufficient to represent arbitrary finite mixtures through iterative composition.
Moreover, its closure under addition and preservation of affine branch structure ensure that symbolic expressions remain interpretable and stable as mixture complexity grows. These properties are intrinsic to the log-domain representation and do not depend on any particular symbolic regression algorithm.
They provide a structural justification for the mixture experiments presented in the main text, without introducing additional modeling assumptions or $n$-ary primitives.

\section{Detailed Experimental Results}
\label{appendix:detailed_results}
\label{appendix:param_shift}

Table~\ref{tab:detailed_recovery} details the recovery outcomes for all benchmark distributions under noiseless and noisy conditions, organized by the four structural categories of SDEBench. While we primarily use $M=50{,}000$, this was increased for complex distributions to ensure robust recovery.

\paragraph{Structural Discovery, Refinement, and Difficulty Spectrum.}
The recorded evolutionary steps indicate the point in the search trace at which the algorithm first identifies the correct symbolic structure, and are therefore distinct from both the fixed search-iteration budget in the implementation configuration and the subsequent parameter refinement stage. We use these step counts as a practical measure of the combinatorial difficulty of the search task. The benchmark families span the four structural classes defined in SDEBench: (i) discrete exponential-family distributions (Poisson, Binomial, Geometric, Neg.-Binomial, Logseries, Hypergeometric, Neg.-Hypergeometric), (ii) power-law and heavy-tailed laws (Zipf, Zipfian, Yule--Simon), (iii) mixed discrete models combining $\log C$- and $\log B$-type structure (Beta-Binomial, Beta-Neg.-Binomial), and (iv) additional structured distributions (Boltzmann, Discrete-Laplace). These families differ substantially in symbolic difficulty: exponential-family distributions (e.g., Geometric, Poisson) are recovered within a few hundred generations, whereas more complex combinatorial families require substantially deeper search---Beta-Binomial requires 12,803 generations and Negative-Hypergeometric requires 42,730. The reported step counts thus provide an empirical view of the difficulty spectrum across structural classes.

\begin{table}[H]
\centering
\scriptsize
\setlength{\tabcolsep}{3pt}
\caption{Comprehensive Symbolic Recovery and Parameter Estimation, organized by structural category. The Steps column marks the generation of structural discovery under noisy conditions; Runtime reports wall-clock time. Within each category, noiseless results (empty Noisy column) precede noisy results.}
\label{tab:detailed_recovery}
\resizebox{\linewidth}{!}{%
\begin{tabular}{L{1.8cm} L{5.4cm} L{2.6cm} L{2.6cm} L{0.8cm} L{1.0cm} L{1.2cm}}
\toprule
\textbf{Inferred Type} & \textbf{Equation} & \textbf{Inferred Params} & \textbf{True Params} & \textbf{Noisy} & \textbf{Steps} & \textbf{Runtime} \\
\midrule
\multicolumn{7}{l}{\textit{(i) Discrete exponential family}} \\
poisson & ((x0 - logF(x0)) + ((x0 * 1.484) - 8.653)) + -3.346 & lam=12.00 & lam=12.0 & & & \\
binomial & (-logF(49.99 - x0) + 130.64) + (-logF(x0) - (x0 * 0.847)) & n=50, p=0.30 & n=50, p=0.30 & & & \\
geometric & (x0 * -0.467) - ((x0 * -0.005) + 0.532) & p=0.37 & p=0.37 & & & \\
negbinomial & ((-logF(x0) + (x0 * -0.356)) + logF(x0 - 48.99)) + -204.76 & r=49.99, p=0.70 & r=50, p=0.70 & & & \\
logseries & (2.605 - ((log(x0) + (x0 * -0.005)) + 1.833)) - x0 & p=0.37 & p=0.37 & & & \\
hypergeometric & (((logC(50.00647, x0 + 10.006) + -64.80) - logC(99.99, x0)) + logC(50.00, x0)) - 0.0008 & N=200, K=80, n=60 & N=200, K=80, n=60 & & & \\
neghypergeometric & logC(60.086, x0 + 0.037) - (131.21 - logC(139.1, x0 + 59.42)) & N=200, K=80, r=60 & N=200, K=80, r=60 & & & \\
poisson & (x0 * 2.485) + (-logF(x0) - 12.01) & lam=12.01 & lam=12.0 & yes & 367 & 1m 19s \\
binomial & ((-logF(x0) - (x0 * 0.845)) + 11.52) + -logF(9.99 - x0) & n=10, p=0.30 & n=10, p=0.30 & yes & 1233 & 4m 05s \\
geometric & (x0 * -0.461) + -0.532 & p=0.37 & p=0.37 & yes & 79 & 29s \\
negbinomial & logF(x0 + 8.99) + (((-24.84 - logF(x0)) + (x0 * -0.356)) - -0.0001) & r=9.99, p=0.70 & r=10, p=0.70 & yes & 1445 & 4m 55s \\
logseries & 0.778 - (log(x0) + x0) & p=0.37 & p=0.37 & yes & 281 & 57s \\
hypergeometric & (logC(60.16, x0) + -131) + logC(138.9, 79.37 - x0) & N=200, K=80, n=61 & N=200, K=80, n=60 & yes & 5019 & 17m 40s \\
neghypergeometric & (logC(80.34, x0) + 83.11) - logC(204.72, x0 + 61.22) & N=205, K=80, r=61 & N=200, K=80, r=60 & yes & 42730 & 149m 30s \\
\midrule
\multicolumn{7}{l}{\textit{(ii) Power-law and heavy-tailed}} \\
zipf & (log(x0) + 0.423) * -1.700 & a=1.70, loc=0 & a=1.70, loc=0 & & & \\
zipfian & ((log(x0) * 0.276) + 0.115) * -6.15 & a=1.7, N=500, loc=0 & a=1.7, N=500, loc=0 & & & \\
yulesimon & logB(2.7, x0 + 0.0) + 0.530628 & rho=1.70 & rho=1.70 & & & \\
zipf & -0.719 - (log(x0) * 1.700) & a=1.70, loc=0 & a=1.70, loc=0 & yes & 286 & 1m 03s \\
zipfian & (log(x0) * -1.700) + -0.710 & a=1.70, N=499 & a=1.7, N=500 & yes & 294 & 1m 01s \\
yulesimon & logB(2.69, x0) + 0.521 & rho=1.69 & rho=1.70 & yes & 261 & 55s \\
\midrule
\multicolumn{7}{l}{\textit{(iii) Mixed discrete models}} \\
betabinomial & ((x0 - logB(101.000656 - x0, 4.0004816)) - logB(1.0003295, x0 + 1.0010041)) + (-22.646421 - x0) & n=100, alpha=1.99, beta=4.99 & n=100, alpha=2.0, beta=5.0 & & & \\
betanegbinomial & ((logB(7.0057716, x0 + 4.994281) - logB(x0 + 1.0000004, 4.00576)) - 3.26e-6) + 2.011 & r=4.99, alpha=5.01, beta=2.00 & r=5, alpha=5, beta=2 & & & \\
betabinomial & (-22.75 - logB(101.08 - x0, 4.02)) - logB(1.009, x0 + 1.02) & n=100, alpha=1.98, beta=4.90 & n=100, alpha=2.0, beta=5.0 & yes & 12803 & 43m 50s \\
betanegbinomial & logB(7.28, x0 + 4.72) + (1.87 - logB(4.28, x0 + 0.99)) & r=6.07, alpha=4.25, beta=2.00 & r=5, alpha=5, beta=2 & yes & 12006 & 41m 20s \\
\midrule
\multicolumn{7}{l}{\textit{(iv) Additional structured distributions}} \\
boltzmann & (x0 * -0.721) + ((x0 * -0.008) - 0.659) & beta=0.73, N=100 & beta=0.73, N=100 & & & \\
dlaplace & (-0.913 - (abs(x0) * -0.048)) - (abs(x0) * 0.898) & a=0.85, loc=0 & a=0.85, loc=0 & & & \\
boltzmann & (x0 * -0.729) + -0.659 & beta=0.73, N=100 & beta=0.73, N=100 & yes & 336 & 1m 14s \\
dlaplace & (abs(x0) * -0.849) + -0.913 & a=0.85, loc=0 & a=0.85, loc=0 & yes & 362 & 1m 12s \\
\bottomrule
\end{tabular}%
}
\end{table}

\subsection{Robustness to Changed Parameter Settings and Supports}
\label{appendix:param_shift_sub}
We evaluate three representative families under substantially changed parameter settings and supports: Poisson, Yule--Simon, and Hypergeometric, covering exponential-family, heavy-tailed, and combinatorial regimes, respectively. SDE recovers the correct family-level symbolic structure with readable closed-form expressions across all settings, though harder configurations can lead to larger numerical error.

\begin{table*}[t]
\centering
\scriptsize
\setlength{\tabcolsep}{5pt}
\caption{Robustness of SDE under changed parameter settings and supports.}
\label{tab:appendix_param_shift}
\begin{tabular}{lp{0.62\linewidth}c}
\toprule
Setting & Recovered Expression & MSE \\
\midrule

\multicolumn{3}{l}{\textbf{Poisson} \quad Target: $\log P(x)=x_0\log\lambda-\lambda-\logF(x_0)$} \\
\cmidrule(lr){1-3}
$\lambda = 2$ &
$1.7049x_0-\logF(x_0)-x_0-2.0374$ &
$0.0064$ \\
$\lambda = 30$ &
$-29.9853+3.4019x_0-\logF(x_0)$ &
$0.0188$ \\

\specialrule{0.12em}{0.08em}{0.06em}
\multicolumn{3}{l}{\textbf{Yule--Simon} \quad Target: $\log P(x)=\log\rho+\log B(x_0,\rho+1)$} \\
\cmidrule(lr){1-3}
$\rho = 0.5$ &
$-0.6375+\log B(1.4978,x_0)$ &
$0.0013$ \\
$\rho = 3.0$ &
$\log B(3.9605,x_0)+1.0386$ &
$0.3446$ \\

\specialrule{0.12em}{0.08em}{0.06em}
\multicolumn{3}{l}{\textbf{Hypergeometric} \quad Target: $\log P(x)=\log C(K,x_0)+\log C(N-K,n-x_0)-\log C(N,n)$} \\
\cmidrule(lr){1-3}
$(N,K,n)=(100,50,40)$ &
$\log C(50.0065,x_0+10.0065)+\log C(50.0060,x_0)-64.7980$ &
$0.0324$ \\
$(N,K,n)=(200,60,100)$ &
$-144.3657+\bigl(\log C(137.2004,104.0455-x_0)+11.2454\bigr)+\log C(60.3952,x_0+2.8689)$ &
$0.1699$ \\
\bottomrule
\end{tabular}
\vskip -0.05in
\end{table*}

\subsection{Extended Composite-Model Recovery Results}
\label{appendix:extra_composite}
We evaluate two further composite distributions under the same finite-sample noisy setting ($M=50{,}000$): a Zero-Inflated Binomial (ZIB) and a mixture of two Poissons. SDE recovers the correct composite symbolic structure and parameters close to the ground truth in both cases, confirming that the framework extends to richer zero-inflated and mixture distributions with minimal modification.

\begin{table*}[t]
\centering
\scriptsize
\setlength{\tabcolsep}{4pt}
\caption{Additional composite-model recovery results ($M=50{,}000$).}
\label{tab:appendix_extra_composite}
\resizebox{\linewidth}{!}{%
\begin{tabular}{l p{0.34\linewidth} l ccc ccc c}
\toprule
Distribution & Recovered Expression (Log-PMF) & Parameters & \multicolumn{3}{c}{Ground Truth} & \multicolumn{3}{c}{Inferred} & MSE \\
\cmidrule(lr){4-6} \cmidrule(lr){7-9}
 & & & 1 & 2 & 3 & 1 & 2 & 3 & \\
\midrule
Zero-Inflated Binomial (ZIB) &
$-1.386+\logaddexp(\log\delta_0(x),$ \newline $\log C(15.079,x)-0.425x-6.967)$ &
$(n,p,\pi)$ &
15 & 0.4 & 0.25 &
15.001 & 0.395 & 0.250 &
0.2179 \\

Mixture of Two Poissons &
$\logaddexp(1.124x-3.413,\ 2.210x-9.950)-\logF(x)-0.017$ &
$(\lambda_1,w,\lambda_2)$ &
3.0 & 0.65 & 9.0 &
3.077 & 0.622 & 9.116 &
0.0664 \\
\bottomrule
\end{tabular}%
}
\vskip -0.05in
\end{table*}

\subsection{Robustness Across Sample Sizes}
\label{appendix:sample_size_robustness}
We evaluate robustness across sample sizes on four families with distinct structural complexity: Yule--Simon, Beta-Binomial, ZIP, and a three-component Binomial mixture. Yule--Simon remains stable down to $M=1{,}000$, degrading to a power-law approximation only at $M=500$. Beta-Binomial is more challenging: recovery is accurate at $M=10{,}000$ and finds a related Beta-function form at $M=5{,}000$, but degrades at smaller sizes. A similar trend holds for ZIP and the Binomial mixture. Together, these results show that simpler targets remain recoverable at small sample sizes, while more combinatorial or composite targets require more data for faithful symbolic recovery.
\begin{table*}[t]
\centering
\scriptsize
\setlength{\tabcolsep}{5pt}
\caption{Additional robustness results across sample sizes.}
\label{tab:appendix_sample_size_robustness}
\begin{tabular}{c c p{0.68\linewidth}}
\toprule
Sample Size ($M$) & MSE & Recovered Expression \\
\midrule

\multicolumn{3}{l}{\textbf{Yule--Simon} \quad $(\rho=1.7,\ K_{\max}=200)$} \\
\cmidrule(lr){1-3}
$10{,}000$ & $0.0021$ &
\begin{tabular}[t]{@{}l@{}}$\log B(2.7100, x_0) + 0.5110$\end{tabular} \\
$5{,}000$ & $0.0153$ &
\begin{tabular}[t]{@{}l@{}}$\log B(x_0, 2.5607)$\end{tabular} \\
$1{,}000$ & $0.0153$ &
\begin{tabular}[t]{@{}l@{}}$\log B(2.5607, x_0)$\end{tabular} \\
$500$ & $0.1338$ &
\begin{tabular}[t]{@{}l@{}}$-2.4169 \log(x_0)$\end{tabular} \\

\specialrule{0.12em}{0.08em}{0.06em}
\multicolumn{3}{l}{\textbf{Beta-Binomial} \quad $(n=100,\ \alpha=2.0,\ \beta=5.0)$} \\
\cmidrule(lr){1-3}
$10{,}000$ & $0.0013$ &
\begin{tabular}[t]{@{}l@{}}$-22.6464 - \log B(1.0003, x_0 + 1.0009) - \log B(101.0007 - x_0, 4.0005)$\end{tabular} \\
$5{,}000$ & $0.0329$ &
\begin{tabular}[t]{@{}l@{}}$\log B(24.2827, 113.2000 - x_0) - \log B(102.6506 - x_0, 23.3376)$\end{tabular} \\
$1{,}000$ & $0.2120$ &
\begin{tabular}[t]{@{}l@{}}$-31.2557 - \log B(7.2884, \log C(161.0127, x_0 + 56.6708))$\end{tabular} \\
$500$ & $0.6720$ &
\begin{tabular}[t]{@{}l@{}}$-15.8934 - \log B(2.8866, 100.2188 - x_0)$\end{tabular} \\

\specialrule{0.12em}{0.08em}{0.06em}
\multicolumn{3}{l}{\textbf{ZIP} \quad $(\pi=0.35,\ \lambda=3.0,\ K_{\max}=100)$} \\
\cmidrule(lr){1-3}
$10{,}000$ & $0.0097$ &
\begin{tabular}[t]{@{}l@{}}$-1.0498 + 1.0986x_0 + \logaddexp(-2.3810, \log\delta_0(x_0)) - \logF(x_0)$\end{tabular} \\
$5{,}000$ & $0.0271$ &
\begin{tabular}[t]{@{}l@{}}$\logaddexp(-3.8617, \log\delta_0(x_0)) - \logF(x_0 - 0.6655) + x_0$\end{tabular} \\
$1{,}000$ & $0.0301$ &
\begin{tabular}[t]{@{}l@{}}$16.0941 - \logaddexp(x_0, 0.3565x_0 + 10.0329) - \logaddexp(x_0, 7.0246)$\end{tabular} \\
$500$ & $0.1128$ &
\begin{tabular}[t]{@{}l@{}}$-0.4537x_0 - \logaddexp(7.8835, x_0) + 6.9262$\end{tabular} \\

\specialrule{0.12em}{0.08em}{0.06em}
\multicolumn{3}{l}{\textbf{Three-component Binomial Mixture} \quad $(n=7,\ \mathbf{p}=\{0.25,0.55,0.8\},\ \mathbf{w}=\{0.5,0.3,0.2\},\ K_{\max}=40)$} \\
\cmidrule(lr){1-3}
$10{,}000$ & $0.0034$ &
\begin{tabular}[t]{@{}l@{}}$\log C(7.3495, x_0) - 14.2495 + \logaddexp(x_0, 3.6969)$\\${}+ \logaddexp(3.2956, -1.1757x_0 + 7.8297)$\end{tabular} \\
$5{,}000$ & $0.0230$ &
\begin{tabular}[t]{@{}l@{}}$\logaddexp(-1.5723x_0 - 2.7054, -8.2355) + \log C(8.4622, x_0) + 0.3359x_0$\end{tabular} \\
$1{,}000$ & $0.1123$ &
\begin{tabular}[t]{@{}l@{}}$\logaddexp(-4.7904, -1.4249x_0) - 2.6877 + \log C(9.6926, x_0)$\end{tabular} \\
$500$ & $0.2340$ &
\begin{tabular}[t]{@{}l@{}}$-15.8208x_0 + x_0 + \log C(9.0117 \times 10^6, x_0) - 3.0000$\end{tabular} \\
\bottomrule
\end{tabular}
\vskip -0.05in
\end{table*}

\begin{table}[t]
\centering
\caption{Geometric $\hat p$ vs.\ $M$ (mean $\pm$ std, 50 reps; true $p=0.30$). SDE recovers the correct log-PMF at all sample sizes, with estimates converging toward the MLE as $M$ grows.}
\label{tab:geom_small_sample}
\scriptsize\setlength{\tabcolsep}{4pt}
\begin{tabular}{ccc}
\toprule
$M$ & SDE & MLE \\
\midrule
50   & $0.26\pm0.05$ & $0.31\pm0.01$ \\
100  & $0.33\pm0.04$ & $0.32\pm0.01$ \\
500  & $0.27\pm0.04$ & $0.30\pm0.01$ \\
1000 & $0.29\pm0.02$ & $0.30\pm0.00$ \\
2000 & $0.29\pm0.01$ & $0.30\pm0.00$ \\
\bottomrule
\end{tabular}
\end{table}

\section{Component-Level Recovery Results for Mixture Models}
\label{app:mixture_additional}
We report detailed component-level parameter recovery results for
mixtures of Binomial distributions with increasing numbers of components.
All components share the same number of trials $n$, while success
probabilities and mixture weights vary.
\paragraph{Two-component mixture.}
Table~\ref{tab:mix2_full} reports component-level recovery results for a mixture
of two binomial distributions.
Both component probabilities and mixture weights are recovered accurately,
with small absolute deviations from the ground-truth values.
Recovered expression:
\scriptsize
\begin{quote}
\texttt{\detokenize{
((logC(5.0257273, x0) + (x0 + logaddexp(x0 * 1.2419002, 3.3747811)))
 - x0 - (x0 * 0.85618645)) - 5.588387
}}
\end{quote}
\normalsize

\begin{table}[ht]
\centering
\scriptsize
\caption{
Component-level parameter recovery for a two-component mixture of binomial distributions.
}
\setlength{\tabcolsep}{6pt}
\begin{tabular}{c c c c c c c}
\toprule
Component & Pred.\ $p$ & GT $p$ & $\Delta p$ & Pred.\ $w$ & GT $w$ & $\Delta w$ \\
\midrule
\#1 & 0.298137 & 0.300000 & $-1.86\times 10^{-3}$ & 0.647595 & 0.650000 & $-2.41\times 10^{-3}$ \\
\#2 & 0.595250 & 0.600000 & $-4.75\times 10^{-3}$ & 0.352405 & 0.350000 & $+2.41\times 10^{-3}$ \\
\bottomrule
\end{tabular}

\label{tab:mix2_full}
\end{table}
\paragraph{Three-component mixture.}
Table~\ref{tab:mix3_full} shows results for a three-component mixture.
The recovered parameters closely match the ground truth for all components,
with moderate degradation compared to the two-component case,
reflecting the increased model complexity.
Recovered expression:
\scriptsize
\begin{quote}
\texttt{\detokenize{
(logC(7.1045327, x0) + ((x0 * 0.117491096) + logaddexp(logaddexp((x0 * -1.2380574) + 8.818689, 4.9220943), (x0 + -0.5984714) * 1.1380522)) + -11.528368
}}
\end{quote}
\normalsize

\begin{table}[ht]
\centering
\scriptsize
\caption{
Component-level parameter recovery for a three-component mixture of binomial distributions.
}
\setlength{\tabcolsep}{6pt}
\begin{tabular}{c c c c c c c}
\toprule
Component & Pred.\ $p$ & GT $p$ & $\Delta p$ & Pred.\ $w$ & GT $w$ & $\Delta w$ \\
\midrule
\#1 & 0.245906 & 0.250000 & $-4.09\times 10^{-3}$ & 0.514311 & 0.500000 & $+1.43\times 10^{-2}$ \\
\#2 & 0.529339 & 0.550000 & $-2.07\times 10^{-2}$ & 0.283124 & 0.300000 & $-1.69\times 10^{-2}$ \\
\#3 & 0.778258 & 0.800000 & $-2.17\times 10^{-2}$ & 0.202565 & 0.200000 & $+2.57\times 10^{-3}$ \\
\bottomrule
\end{tabular}
\label{tab:mix3_full}
\end{table}
\paragraph{Four-component mixture.}
Table~\ref{tab:mix4_full} presents results for a mixture with four components.
While the symbolic structure corresponding to a valid mixture is recovered,
parameter estimation errors increase for some components,
illustrating the growing difficulty of disentangling multiple overlapping modes.
Recovered expression:
\scriptsize
\begin{quote}
\texttt{\detokenize{
(x0 + logC(7.2584515, x0)) + (((((x0 + 0.06383211) * -0.77610856) + -6.0938134) + logaddexp(-6.7171907, (x0 + 0.035176843) * -1.3297195)) + logaddexp(3.293205, x0 * 1.1077892))
}}
\end{quote}
\normalsize

\begin{table}[ht]
\centering
\scriptsize
\caption{
Component-level parameter recovery for a four-component mixture of binomial distributions.
}
\setlength{\tabcolsep}{6pt}
\begin{tabular}{c c c c c c c}
\toprule
Component & Pred.\ $p$ & GT $p$ & $\Delta p$ & Pred.\ $w$ & GT $w$ & $\Delta w$ \\
\midrule
\#1 & 0.248650 & 0.250000 & $-1.35\times 10^{-3}$ & 0.437240 & 0.400000 & $+3.72\times 10^{-2}$ \\
\#2 & 0.500490 & 0.450000 & $+5.05\times 10^{-2}$ & 0.314350 & 0.250000 & $+6.44\times 10^{-2}$ \\
\#3 & 0.555740 & 0.650000 & $-9.43\times 10^{-2}$ & 0.025130 & 0.200000 & $-1.75\times 10^{-1}$ \\
\#4 & 0.791120 & 0.850000 & $-5.89\times 10^{-2}$ & 0.223280 & 0.150000 & $+7.33\times 10^{-2}$ \\
\bottomrule
\end{tabular}
\label{tab:mix4_full}
\end{table}
\paragraph{Six-component mixture.}
Table~\ref{tab:mix6_full} reports results for a more challenging six-component mixture.
Although the recovered expression still represents a valid mixture structure,
parameter recovery degrades substantially for several components,
highlighting the practical limits of symbolic recovery as mixture complexity increases. Recovered expression:
\scriptsize
\begin{quote}
\texttt{\detokenize{
(logaddexp(logaddexp((x0 * -12.915779) + 2.8190386, 6.7233334), (x0 + 0.053147778) * 1.2876362) + (logC(8.071398, x0) + logaddexp(x0 * 1.1566708, 2.411895))) + (((x0 + -0.18561293) * -1.2707464) + -12.915763)
}}
\end{quote}
\normalsize

\begin{table}[ht]
\centering
\scriptsize
\caption{
Component-level parameter recovery for a six-component mixture of binomial distributions.
}
\setlength{\tabcolsep}{6pt}
\begin{tabular}{c c c c c c c}
\toprule
Component & Pred.\ $p$ & GT $p$ & $\Delta p$ & Pred.\ $w$ & GT $w$ & $\Delta w$ \\
\midrule
\#1 & 0.00000069 & 0.200000 & $-2.00\times 10^{-1}$ & 0.00123 & 0.100000 & $-9.88\times 10^{-2}$ \\
\#2 & 0.00000219 & 0.350000 & $-3.50\times 10^{-1}$ & 0.00011 & 0.200000 & $-1.999\times 10^{-1}$ \\
\#3 & 0.219130 & 0.500000 & $-2.81\times 10^{-1}$ & 0.34350 & 0.150000 & $+1.94\times 10^{-1}$ \\
\#4 & 0.471510 & 0.650000 & $-1.78\times 10^{-1}$ & 0.47344 & 0.300000 & $+1.73\times 10^{-1}$ \\
\#5 & 0.504220 & 0.800000 & $-2.96\times 10^{-1}$ & 0.01064 & 0.050000 & $-3.94\times 10^{-2}$ \\
\#6 & 0.763790 & 0.900000 & $-1.36\times 10^{-1}$ & 0.17108 & 0.200000 & $-2.89\times 10^{-2}$ \\
\bottomrule
\end{tabular}
\label{tab:mix6_full}
\end{table}

\section{Zero-Inflated Model Details}
\subsection{Unified formulation and operators}

As described in the main text, a zero-inflated distribution combines a discrete atomic mass at zero with a baseline count distribution. In the log-domain, all zero-inflated families considered in this work can be written in the unified form:
\begin{equation}
\log p(x) = \operatorname{logaddexp}\!\Big( \log \pi + \log \delta_0(x), \log(1-\pi) + \log g(x;\theta) \Big),
\end{equation}
where $\log \delta_0(x)=0$ if $x=0$ and is assigned a large negative constant otherwise.

\subsection{Description of Diagnostic Plots}
\begin{figure}[ht]
    \centering
    \begin{subfigure}[b]{0.85\linewidth}
        \centering
        \includegraphics[width=\linewidth]{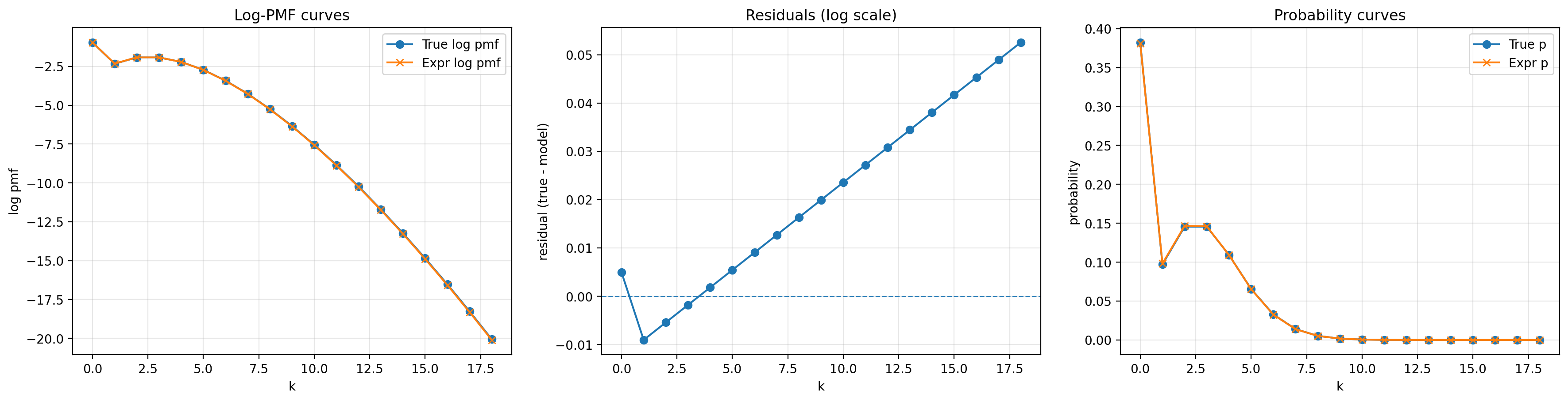}
        \caption{ZIP: recovered model accurately recovers the Poisson baseline and zero-inflation atom.}
        \label{fig:zip-app}
    \end{subfigure}
    \\[1ex]
    \begin{subfigure}[b]{0.85\linewidth}
        \centering
        \includegraphics[width=\linewidth]{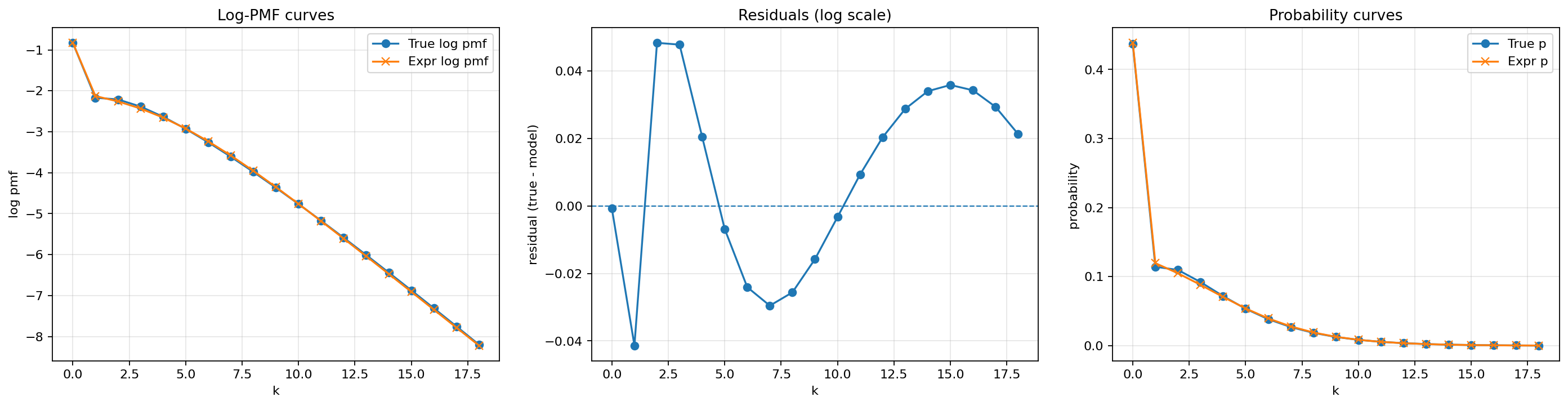}
        \caption{ZINB: recovery captures the heavy-tailed baseline and excess-zero mechanism.}
        \label{fig:zinb-app}
    \end{subfigure}
    \\[1ex]
    \begin{subfigure}[b]{0.85\linewidth}
        \centering
        \includegraphics[width=\linewidth]{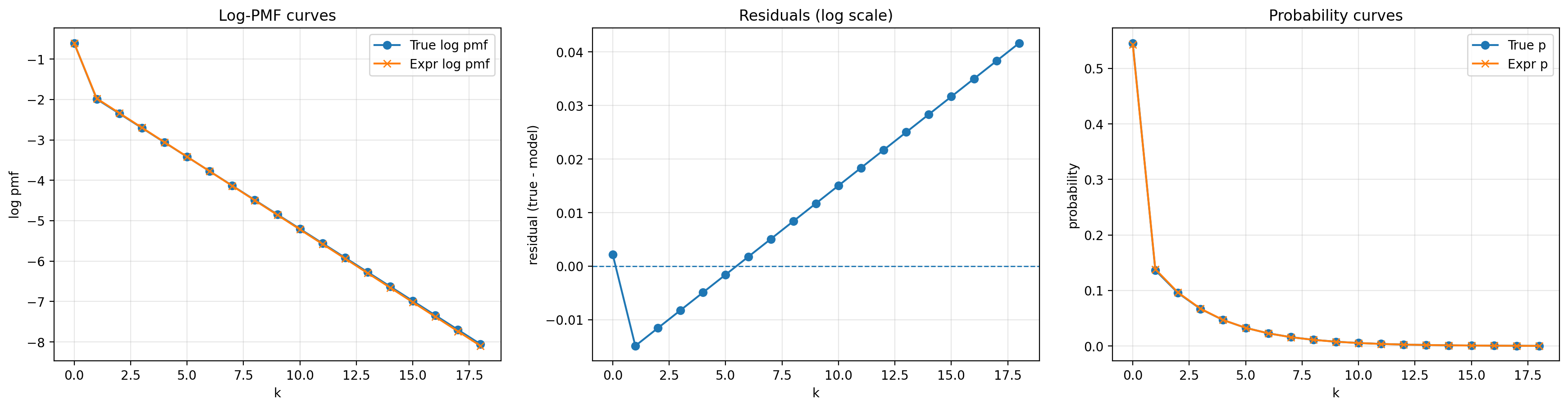}
        \caption{ZIG: concise symbolic form reflects geometric decay and zero-inflation.}
        \label{fig:zig-app}
    \end{subfigure}
    \caption{Diagnostic plots for zero-inflated models (ZIP, ZINB, ZIG). Each panel shows PMF fit, residuals, and reconstruction accuracy.}
    \label{fig:zi-diagnostics}
\end{figure}

To evaluate the fidelity of the recovered symbolic expressions, we provide three-panel diagnostic plots for each distribution family (Figures~\ref{fig:zip-app}--\ref{fig:zig-app}). These plots comprise:
\begin{itemize}
    \item \textbf{Log-PMF curves}: A direct comparison between the ground-truth log-PMF and the values produced by the recovered symbolic formula, demonstrating the structural fit in the log-domain.
    \item \textbf{Residuals (log scale)}: The point-wise difference (True $-$ Model) across the support $x$, used to quantify local approximation errors.
    \item \textbf{Probability curves}: The resulting probability mass function (PMF) after applying a softmax transformation to the log-domain expressions, illustrating the recovered distribution and the zero-inflation effect in the natural probability space.
\end{itemize}

\subsection{Zero-Inflated Poisson (ZIP)}
For the Zero-Inflated Poisson distribution, the symbolic regression recovers the following log-domain expression:
\begin{equation}
y(x) = (x \cdot 1.0949) + \text{logaddexp} ( -2.3616 + \log \delta_0(x), -1.0564 - \text{logF}(x) )
\end{equation}

\subsection{Zero-Inflated Negative Binomial (ZINB)}
The Zero-Inflated Negative Binomial introduces an additional shape parameter in the baseline distribution. The recovered symbolic expression maintains the unified zero-inflated structure:
\begin{equation}
\log \delta_0(x) + ( \text{logaddexp}( -0.3135 - \log \delta_0(x), 0.5358 ) - \text{logaddexp}( 0.4392 \cdot x, 1.5226 ) )
\end{equation}

\subsection{Zero-Inflated Geometric (ZIG)}
The geometric baseline yields a highly compact symbolic representation. The recovered expression explicitly encodes the zero-inflation mechanism:
\begin{equation} \text{logaddexp}(\log \delta_0(x), -0.5530) + 1.9112 - (0.36x + 2.9747) \end{equation}

\section{Baseline Comparison: Extended Results}
\paragraph{Comprehensive Parameter Estimation Results.}
Table~\ref{tab:parameter_results} presents the complete set of inferred parameters for all evaluated distributions, extending the summary in the main text. SDE consistently recovers parameters with high fidelity, even in challenging mixture and zero-inflated settings. MaxErr percentages are averaged over 100 independent seeds at $M=50{,}000$; the table below shows a representative single run.

\begin{table*}[t]
\centering
\caption{Comparison of parameter estimation results. Symbols ``--'' indicate parameters that are not applicable (e.g., $n$ is fixed and not estimated) or for which the method does not apply to that family.}
\label{tab:parameter_results}
\vskip 0.15in
\scriptsize
\resizebox{\linewidth}{!}{%
\begin{tabular}{lccccc}
\toprule
\textbf{Distribution} & \textbf{Parameters} & \textbf{True Values} & \textbf{MoM Est.} & \textbf{MLE/EM Est.} & \textbf{SDE (Ours)} \\
\midrule
Poisson & $\lambda$ & 12.0 & 12.00 & 12.00 & 12.01 \\
Binomial & $n, p$ & 10, 0.3 & --, 0.30 & --, 0.30 & \textbf{10, 0.30} \\
Geometric & $p$ & 0.3 & 0.30 & 0.30 & 0.30 \\
NegBinomial & $r, p$ & 10.0, 0.7 & 10.03, 0.70 & 9.99, 0.70 & \textbf{9.99, 0.70} \\
BetaBinomial & $n, \alpha, \beta$ & 100, 2.0, 5.0 & --, 1.84, 4.61 & --, 2.00, 5.00 & 100, 1.98, 4.90 \\
\midrule
ZIP & $\pi, \lambda$ & 0.35, 3.0 & -- & 0.35, 3.00 & 0.35, 2.99 \\
ZIG & $\pi, p$ & 0.35, 0.3 & -- & 0.35, 0.30 & 0.34, 0.30 \\
ZINB & $\pi, r, p$ & 0.35, 2.2, 0.4 & -- & 0.35, 2.20, 0.40 & 0.35, 2.20, 0.40 \\
\midrule
Binomial mixture & $n$ & 7 & -- & 7.00 & 7.00 \\
& $P$ & $[0.25, 0.55, 0.80]$ & -- & $[0.24, 0.55, 0.79]$ & $[0.25, 0.53, 0.78]$ \\
& $W$ & $[0.50, 0.30, 0.20]$ & -- & $[0.50, 0.30, 0.21]$ & $[0.51, 0.28, 0.20]$ \\
\bottomrule
\end{tabular}%
}
\vskip -0.1in
\end{table*}

\paragraph{Controlled experiment with PySR under matched operator set.}
\label{appendix:pysr_controlled}
We evaluate PySR under the same noisy setting with the same domain-specific operator set and complexity budget as SDE, across three representative families: Zipf, Binomial, and Beta-Binomial (Table~\ref{tab:pysr_controlled}).

For the simple Zipf case, PySR recovers a reasonable logarithmic form. For more structured families such as Binomial and Beta-Binomial, PySR does not recover the expected combinatorial form even under the same operator set and complexity budget; the returned expressions are ad hoc functional fits rather than symbolic PMFs. SDE consistently recovers the expected factorial/combinatorial structure with lower error. The performance gap therefore reflects the validity-aware symbolic PMF search, not merely a difference in operator vocabulary.

\begin{table*}[t]
\centering
\scriptsize
\setlength{\tabcolsep}{3pt}
\renewcommand{\arraystretch}{1.12}
\caption{PySR vs.\ SDE under a matched operator set and complexity budget.}
\label{tab:pysr_controlled}
\begin{tabularx}{\textwidth}{@{}>{\raggedright\arraybackslash}p{3.2cm} c >{\raggedright\arraybackslash}X c >{\raggedright\arraybackslash}X@{}}
\toprule
Distribution & PySR MSE & PySR Expression & SDE MSE & SDE Expression \\
\midrule
Zipf \\[-2pt] \tiny$(a=1.50,\ N=200)$
& $0.015$
& $-0.891 - 1.506 \log(x_0)$
& $0.002$
& $-1.500 \log(x_0) - 0.908$ \\[4pt]

Binomial \\[-2pt] \tiny$(n=10,\ p=0.30)$
& $0.010$
& $-3.604 - x_0(x_0^{0.515} - 2.540)$
& $0.001$
& $-\log F(x_0) - 0.845x_0 + 11.520 - \log F(9.990 - x_0)$ \\[4pt]

Beta-Binomial \\[-2pt] \tiny$(n=100,\ \alpha=2.0,\ \beta=5.0)$
& $0.048$
& $-4.748 - (0.001x_0 - 0.071)(x_0 - \sin(x_0))$
& $0.003$
& $-22.750 - \log B(101.080 - x_0, 4.020) - \log B(1.009, x_0 + 1.020)$ \\
\bottomrule
\end{tabularx}
\vskip -0.05in
\end{table*}

\paragraph{Enumeration baseline and reduced-grammar controls.}
\label{appendix:enumeration}
To assess the roles of search design and operator vocabulary separately, we conduct three controlled experiments on Poisson$(\lambda=10)$. First, we run SDE with uniform operator costs and no structural constraints over the full grammar. Second, we run the same ablated SDE over a reduced Poisson-relevant grammar $\{\logB,\logF,*,+,-,\log,\exp,\wedge\}$. Third, under this reduced grammar, we implement an exhaustive-enumeration baseline followed by Sure Independence Screening (SIS) and LASSO.

Under the ablated full grammar, SDE does not recover the correct Poisson structure, but instead returns a structurally incorrect yet numerically reasonable expression with MSE $=0.0042$. In contrast, under the same ablated setting but with the reduced grammar above, SDE recovers the correct Poisson log-PMF with MSE $=3.2\times10^{-5}$. Even without the complexity-profile design or structural restrictions, the evolutionary search identifies the correct law once the search space is sufficiently focused.

For the enumeration baseline, we assign unit cost to all leaves and operators and enumerate expressions up to oracle complexity 8. Even under the reduced grammar, this already yields 400,376 candidate expressions. Direct LASSO on the full enumerated library is infeasible, so we use a two-stage pipeline: we first compute SIS scores for the enumerated features, retain the top-200 screened expressions, and then fit LASSO on this reduced design matrix. In practice, only 136,004 of the 400,376 enumerated expressions can be fit successfully, while 264,372 fail before yielding usable screened features. Running LASSO on the top-200 SIS-selected expressions still does not recover a compact symbolic law, but instead returns a sparse combination with 37 nonzero terms (28 even under the oracle subset), with runtime 2,869s. By contrast, SDE under the same reduced grammar recovers the correct Poisson log-PMF in 120s. One-shot enumeration is thus computationally burdensome and does not recover a single interpretable expression comparable to SDE.

\begin{figure*}[ht]
    \centering
    \begin{subfigure}{\textwidth}
        \centering
        \includegraphics[width=0.9\textwidth]{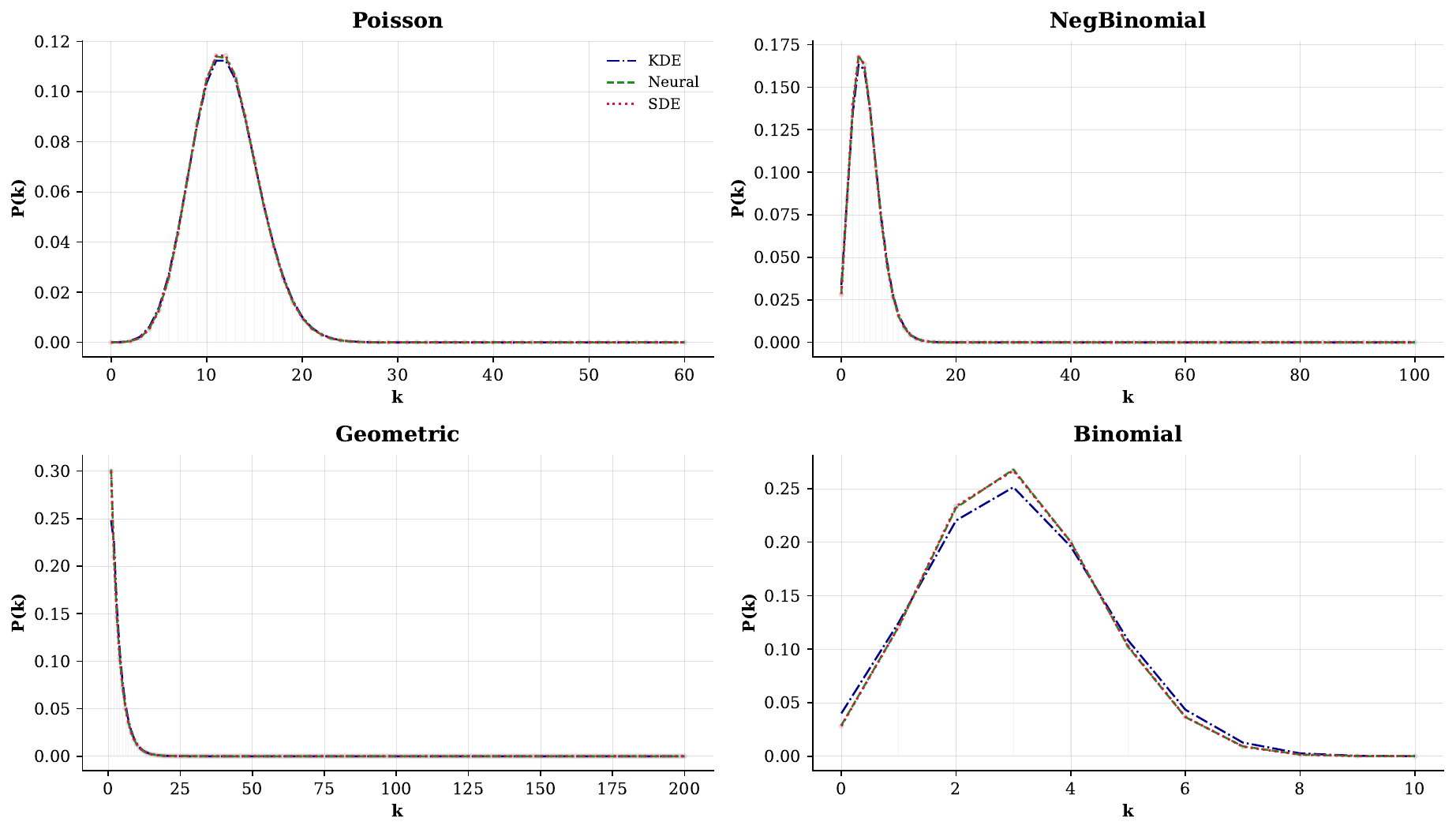}
        \vspace{1mm}
    \end{subfigure}

    \begin{subfigure}{\textwidth}
        \centering
        \includegraphics[width=0.9\textwidth]{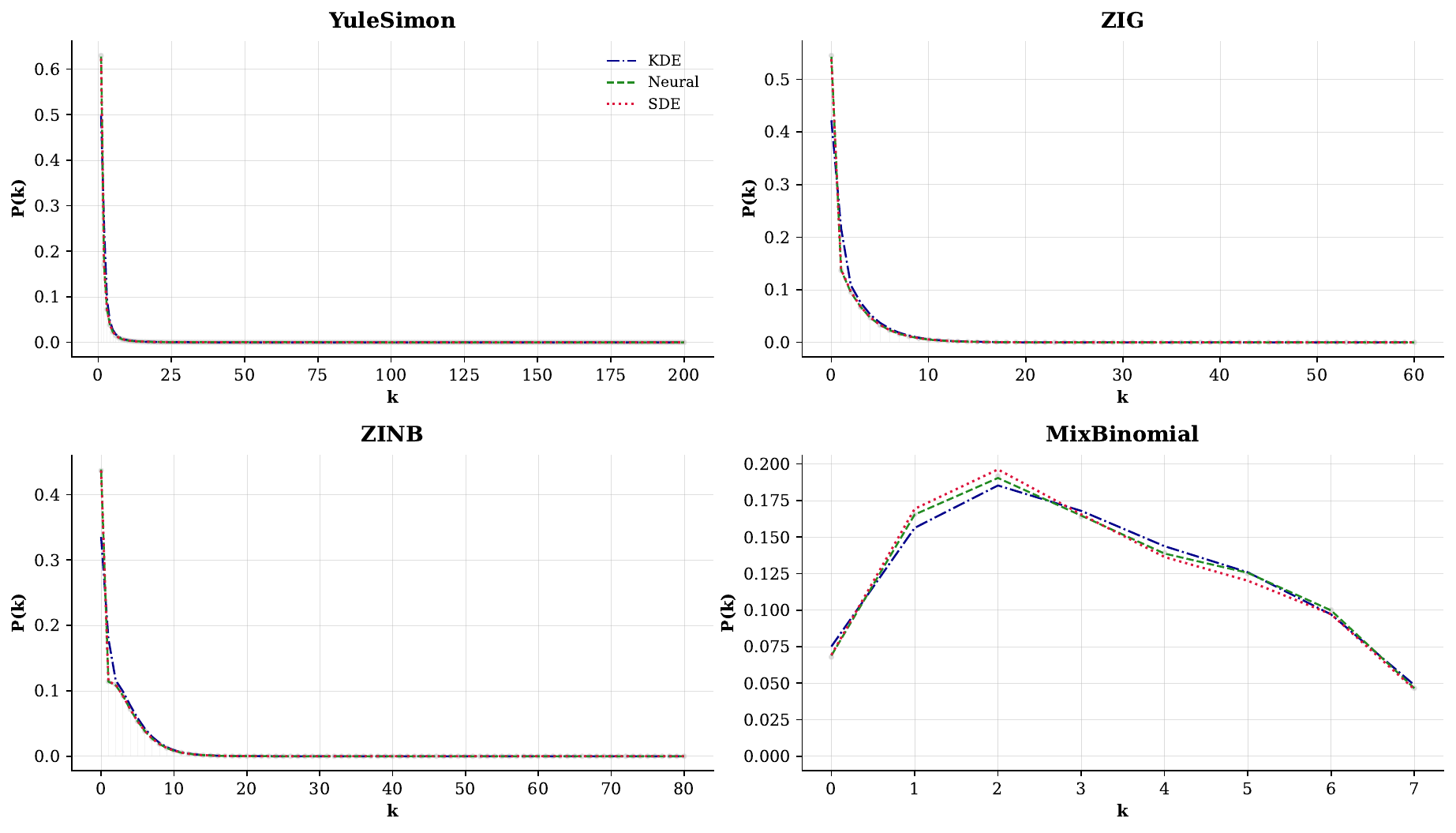}
        \vspace{1mm}
    \end{subfigure}

    \caption{Comparison of PMF estimation results across the extended benchmark. Beta-Binomial and ZIP comparisons (including KDE and Pyro) are shown in Figure~\ref{fig:kde_pyro_comparison} in the main text.}
    \label{fig:appendix_visualizations}
\end{figure*}

\paragraph{Comprehensive Black-Box Density Estimation Results.}
\label{app:Black-Box Density Estimation Results}
We provide recovery plots for the benchmark suite. Figure~\ref{fig:appendix_visualizations} presents the probability mass function (PMF) comparisons for classical distributions (Poisson, Negative Binomial, Geometric, Binomial, Yule--Simon) and complex variants, including zero-inflated and overdispersed models (ZIP, ZIG, ZINB, Beta-Binomial, and Binomial mixture).

For nonparametric baselines such as KDE and Pyro, which lack an explicit analytical form, performance is assessed through visual comparison of the recovered PMFs. While these methods can achieve reasonable empirical fits, they often fail to recover the underlying mathematical structure. For example, KDE exhibits visible inaccuracies near zero in the ZIP distribution. Meanwhile, KDE and Pyro produce non-smooth fluctuations, as seen in the Beta-Binomial plot, indicating a tendency to overfit local sampling noise. In contrast, SDE closely matches the benchmark PMFs with smooth closed-form expressions across the full suite.

\subsection{PBMC Case Study Details}
\label{appendix:pbmc_details}

We use the PBMC3k scRNA-seq dataset \citep{Zheng2017-zs} and focus on a single
gene (Gene 4046), selected at random among genes with sufficient expression counts to support a non-trivial count distribution. Each cell contributes one nonnegative integer
UMI count for this gene, yielding a one-dimensional empirical count
distribution across cells.

To reduce the influence of extreme-quality cells and outliers, we filter cells
by library size (total UMI count per cell) and retain only those within the
$[20\%,98\%]$ quantile range. After filtering, we compute the empirical
distribution of Gene 4046 counts on a finite support of 82 values
($k=0,\dots,81$), and the total observed transcript count for this gene is
2{,}107.

We then construct empirical log-frequency targets and apply the same
noise-aware fitting procedure described in
Section~\ref{sub:smoothing}, using the identical
hyperparameters for smoothing, masking, and weighting as in our implementation.

%% file: references.bib
@book{bernardo1994bayesian,
  author    = {Bernardo, Jos{\'e} M. and Smith, Adrian F. M.},
  title     = {Bayesian Theory},
  publisher = {John Wiley \& Sons},
  address   = {Chichester},
  year      = {1994}
}

@article{li2020comparing,
  author  = {Li, Meng and Dunson, David B.},
  title   = {Comparing and Weighting Imperfect Models Using {D}-Probabilities},
  journal = {Journal of the American Statistical Association},
  year    = {2020},
  volume  = {115},
  number  = {531},
  pages   = {1349--1360},
}

@inproceedings{akaike1973information,
  author    = {Akaike, Hirotugu},
  title     = {Information Theory and an Extension of the Maximum Likelihood Principle},
  booktitle = {Second International Symposium on Information Theory},
  editor    = {Petrov, B. N. and Csaki, F.},
  pages     = {267--281},
  publisher = {Akad{\'e}miai Kiad{\'o}},
  address   = {Budapest},
  year      = {1973}
}

@article{akaike1974new,
  author  = {Akaike, Hirotugu},
  title   = {A New Look at the Statistical Model Identification},
  journal = {IEEE Transactions on Automatic Control},
  year    = {1974},
  volume  = {19},
  number  = {6},
  pages   = {716--723}
}

@article{geisser1979predictive,
  author  = {Geisser, Seymour and Eddy, William F.},
  title   = {A Predictive Approach to Model Selection},
  journal = {Journal of the American Statistical Association},
  year    = {1979},
  volume  = {74},
  number  = {365},
  pages   = {153--160}
}

@article{aitkin1991posterior,
  author  = {Aitkin, Murray},
  title   = {Posterior {Bayes} Factors},
  journal = {Journal of the Royal Statistical Society. Series B (Methodological)},
  year    = {1991},
  volume  = {53},
  number  = {1},
  pages   = {111--142}
}

@article{gelfand1994bayesian,
  author  = {Gelfand, Alan E. and Dey, Dipak K.},
  title   = {Bayesian Model Choice: Asymptotics and Exact Calculations},
  journal = {Journal of the Royal Statistical Society. Series B (Methodological)},
  year    = {1994},
  pages   = {501--514}
}

@book{claeskens2008model,
  author    = {Claeskens, Gerda and Hjort, Nils Lid},
  title     = {Model Selection and Model Averaging},
  publisher = {Cambridge University Press},
  series    = {Cambridge Series in Statistical and Probabilistic Mathematics},
  year      = {2008}
}

@article{hoeting1999bayesian,
  author  = {Hoeting, Jennifer A. and Madigan, David and Raftery, Adrian E. and Volinsky, Chris T.},
  title   = {{Bayesian} Model Averaging: A Tutorial},
  journal = {Statistical Science},
  year    = {1999},
  volume  = {14},
  number  = {4},
  pages   = {382--401}
}

@incollection{clyde2013bayesian,
  author    = {Clyde, Merlise A. and Iversen, Edwin S.},
  title     = {Bayesian Model Averaging in the {M}-Open Framework},
  booktitle = {Bayesian Theory and Applications},
  editor    = {Damien, Paul and Dellaportas, Petros and Polson, Nicholas G. and Stephens, David A.},
  pages     = {483--498},
  publisher = {Oxford University Press},
  year      = {2013}
}

@InProceedings{pan_sr,
  title     = {Ab Initio Nonparametric Variable Selection for Scalable Symbolic Regression with Large $p$},
  author    = {Ye, Shengbin and Li, Meng},
  booktitle = {Proceedings of the 42nd International Conference on Machine Learning},
  pages     = {72041--72062},
  year      = {2025},
  editor    = {Singh, Aarti and Fazel, Maryam and Hsu, Daniel and Lacoste-Julien, Simon and Berkenkamp, Felix and Maharaj, Tegan and Wagstaff, Kiri and Zhu, Jerry},
  volume    = {267},
  series    = {Proceedings of Machine Learning Research},
  publisher = {PMLR}
}

@article{cochran1954,
  author  = {Cochran, William G.},
  journal = {Biometrics},
  number  = {4},
  pages   = {417--451},
  title   = {Some {M}ethods for {S}trengthening the {C}ommon $\chi^2$ {T}ests},
  volume  = {10},
  year    = {1954}
}

@book{agresti2002categorical,
  author    = {Agresti, Alan},
  title     = {Categorical Data Analysis},
  year      = {2002},
  publisher = {John Wiley \& Sons},
  address   = {New York},
  edition   = {2nd}
}

@book{feller1968introduction,
  author    = {Feller, William},
  title     = {An Introduction to Probability Theory and Its Applications},
  year      = {1968},
  volume    = {1},
  edition   = {3rd},
  publisher = {John Wiley \& Sons},
  address   = {New York}
}

@book{johnson1992univariate,
  author    = {Johnson, N. Lloyd and Kemp, Adrienne W. and Kotz, Samuel},
  title     = {Univariate Discrete Distributions},
  edition   = {2nd},
  publisher = {John Wiley \& Sons},
  address   = {New York},
  year      = {1992}
}

@article{2020SciPy-NMeth,
  author  = {Virtanen, Pauli and Gommers, Ralf and Oliphant, Travis E. and
             Haberland, Matt and Reddy, Tyler and Cournapeau, David and
             Burovski, Evgeni and Peterson, Pearu and Weckesser, Warren and
             Bright, Jonathan and {van der Walt}, St{\'e}fan J. and
             Brett, Matthew and Wilson, Joshua and Millman, K. Jarrod and
             Mayorov, Nikolay and Nelson, Andrew R. J. and Jones, Eric and
             Kern, Robert and Larson, Eric and Carey, C.~J. and
             Polat, {\.I}lhan and Feng, Yu and Moore, Eric W. and
             {VanderPlas}, Jake and Laxalde, Denis and Perktold, Josef and
             Cimrman, Robert and Henriksen, Ian and Quintero, E. A. and
             Harris, Charles R. and Archibald, Anne M. and
             Ribeiro, Ant{\^o}nio H. and Pedregosa, Fabian and
             {van Mulbregt}, Paul and {SciPy 1.0 Contributors}},
  title   = {{{SciPy} 1.0: Fundamental Algorithms for Scientific
             Computing in Python}},
  journal = {Nature Methods},
  year    = {2020},
  volume  = {17},
  pages   = {261--272}
}

@book{casella2024statistical,
  author    = {Casella, George and Berger, Roger L.},
  title     = {Statistical Inference},
  edition   = {2nd},
  year      = {2024},
  publisher = {Chapman and Hall/CRC}
}

@article{BRENCE2021107077,
  title   = {Probabilistic grammars for equation discovery},
  journal = {Knowledge-Based Systems},
  volume  = {224},
  pages   = {107077},
  year    = {2021},
  author  = {Brence, Jure and Todorovski, Ljup{\v{c}}o and D{\v{z}}eroski, Sa{\v{s}}o}
}

@misc{schneider2024,
  title        = {Probabilistic Regular Tree Priors for Scientific Symbolic Reasoning},
  author       = {Schneider, Tim and Totounferoush, Amin and Nowak, Wolfgang and Staab, Steffen},
  year         = {2024},
  eprint       = {2306.08506},
  archivePrefix = {arXiv},
  note   = {arXiv:2306.08506},
  primaryClass = {cs.LG}
}

@inproceedings{PGE,
  author    = {Worm, Tony and Chiu, Kenneth},
  title     = {Prioritized grammar enumeration: symbolic regression by dynamic programming},
  year      = {2013},
  publisher = {Association for Computing Machinery},
  address   = {New York, NY, USA},
  booktitle = {Proceedings of the 15th Annual Conference on Genetic and Evolutionary Computation},
  pages     = {1021--1028},
  series    = {GECCO '13}
}

@article{Bartlett2020ExhaustiveSR,
  author  = {Bartlett, Deaglan J. and Desmond, Harry and Ferreira, Pedro G.},
  journal = {IEEE Transactions on Evolutionary Computation},
  title   = {Exhaustive Symbolic Regression},
  year    = {2024},
  volume  = {28},
  number  = {4},
  pages   = {950--964}
}

@article{Udrescu2020AIFeynman,
  author  = {Udrescu, Silviu-Marian and Tegmark, Max},
  title   = {{AI} {Feynman}: A physics-inspired method for symbolic regression},
  journal = {Science Advances},
  volume  = {6},
  number  = {16},
  pages   = {eaay2631},
  year    = {2020}
}

@inproceedings{udrescu2020aifeynman20paretooptimal,
  author    = {Udrescu, Silviu-Marian and Tan, Andrew and Feng, Jiahai and Neto, Orisvaldo and Wu, Tailin and Tegmark, Max},
  booktitle = {Advances in Neural Information Processing Systems},
  editor    = {H. Larochelle and M. Ranzato and R. Hadsell and M.F. Balcan and H. Lin},
  pages     = {4860--4871},
  publisher = {Curran Associates, Inc.},
  title     = {{AI} {Feynman} 2.0: Pareto-optimal symbolic regression exploiting graph modularity},
  volume    = {33},
  year      = {2020}
}

@article{Liu2020,
  author  = {Liu, Chun-Yen and Zhang, Shijia and Martinez, Daniel and Li, Meng and Senftle, Thomas P.},
  title   = {Using statistical learning to predict interactions between single metal atoms and modified {MgO}(100) supports},
  journal = {npj Computational Materials},
  year    = {2020},
  volume  = {6},
  number  = {1},
  pages   = {102}
}

@article{Liu2022-kh,
  title   = {A rapid feature selection method for catalyst design: Iterative {Bayesian} additive regression trees ({iBART})},
  author  = {Liu, Chun-Yen and Ye, Shengbin and Li, Meng and Senftle, Thomas P.},
  journal = {Journal of Chemical Physics},
  volume  = {156},
  number  = {16},
  pages   = {164105},
  year    = {2022}
}

@incollection{Smits2005,
  author    = {Smits, Guido F. and Kotanchek, Mark},
  editor    = {O'Reilly, Una-May and Yu, Tina and Riolo, Rick and Worzel, Bill},
  title     = {Pareto-Front Exploitation in Symbolic Regression},
  booktitle = {Genetic Programming Theory and Practice II},
  year      = {2005},
  publisher = {Springer US},
  address   = {Boston, MA},
  pages     = {283--299}
}

@article{Ye2024-og,
  title   = {Operator-induced structural variable selection for identifying materials genes},
  author  = {Ye, Shengbin and Senftle, Thomas P. and Li, Meng},
  journal = {Journal of the American Statistical Association},
  volume  = {119},
  number  = {545},
  pages   = {81--94},
  year    = {2024}
}

@misc{austel2020,
  title        = {Symbolic Regression using Mixed-Integer Nonlinear Optimization},
  author       = {Austel, Vernon and Cornelio, Cristina and Dash, Sanjeeb and Goncalves, Joao and Horesh, Lior and Josephson, Tyler and Megiddo, Nimrod},
  year         = {2020},
  eprint       = {2006.06813},
  archivePrefix = {arXiv},
  note   = {arXiv:2006.06813},
  primaryClass = {cs.LG}
}

@misc{kim2021,
  title        = {Learning Symbolic Expressions: Mixed-Integer Formulations, Cuts, and Heuristics},
  author       = {Kim, Jongeun and Leyffer, Sven and Balaprakash, Prasanna},
  year         = {2021},
  eprint       = {2102.08351},
  archivePrefix = {arXiv},
  note   = {arXiv:2102.08351},
  primaryClass = {math.OC}
}

@article{NEUMANN2020123412,
  title   = {A new formulation for symbolic regression to identify physico-chemical laws from experimental data},
  journal = {Chemical Engineering Journal},
  volume  = {387},
  pages   = {123412},
  year    = {2020},
  author  = {Neumann, Pascal and Cao, Liwei and Russo, Danilo and Vassiliadis, Vassilios S. and Lapkin, Alexei A.}
}

@incollection{McConaghy2011,
  author    = {McConaghy, Trent},
  editor    = {Riolo, Rick and Vladislavleva, Ekaterina and Moore, Jason H.},
  title     = {FFX: Fast, Scalable, Deterministic Symbolic Regression Technology},
  booktitle = {Genetic Programming Theory and Practice IX},
  year      = {2011},
  publisher = {Springer New York},
  address   = {New York, NY},
  pages     = {235--260}
}

@article{BRUNTON2016710,
  title   = {Sparse Identification of Nonlinear Dynamics with Control ({SINDyC})},
  journal = {IFAC-PapersOnLine},
  volume  = {49},
  number  = {18},
  pages   = {710--715},
  year    = {2016},
  author  = {Brunton, Steven L. and Proctor, Joshua L. and Kutz, J. Nathan}
}

@article{Champion_2019,
  title   = {Data-driven discovery of coordinates and governing equations},
  volume  = {116},
  number  = {45},
  journal = {Proceedings of the National Academy of Sciences},
  author  = {Champion, Kathleen and Lusch, Bethany and Kutz, J. Nathan and Brunton, Steven L.},
  year    = {2019},
  pages   = {22445--22451}
}

@Book{Koza1992,
  author    = {Koza, John R.},
  title     = {Genetic Programming: On the Programming of Computers by Means of Natural Selection},
  year      = {1992},
  publisher = {MIT Press},
  address   = {Cambridge, MA}
}

@book{Poli2008,
  author = {Poli, Riccardo and Langdon, William B. and McPhee, Nicholas F.},
  year   = {2008},
  title  = {A Field Guide to Genetic Programming},
  publisher = {Lulu Press}
}

@article{Vladislavleva2009,
  author  = {Vladislavleva, Ekaterina and Smits, Guido and den Hertog, Dick},
  year    = {2009},
  pages   = {333--349},
  title   = {Order of Nonlinearity as a Complexity Measure for Models Generated by Symbolic Regression via Pareto Genetic Programming},
  volume  = {13},
  journal = {IEEE Transactions on Evolutionary Computation}
}

@article{Uy2011,
  author    = {Uy, Nguyen Quang and Hoai, Nguyen Xuan and O'Neill, Michael and McKay, R. I. and Galv\'{a}n-L\'{o}pez, Edgar},
  title     = {Semantically-based crossover in genetic programming: application to real-valued symbolic regression},
  year      = {2011},
  volume    = {12},
  number    = {2},
  journal   = {Genetic Programming and Evolvable Machines},
  pages     = {91--119}
}

@incollection{Kronberger2019,
  author    = {Kronberger, Gabriel and Kammerer, Lukas and Burlacu, Bogdan and Winkler, Stephan M. and Kommenda, Michael and Affenzeller, Michael},
  editor    = {Banzhaf, Wolfgang and Spector, Lee and Sheneman, Leigh},
  title     = {Cluster Analysis of a Symbolic Regression Search Space},
  booktitle = {Genetic Programming Theory and Practice XVI},
  year      = {2019},
  publisher = {Springer International Publishing},
  address   = {Cham},
  pages     = {85--102}
}

@misc{PYSR,
  title        = {{PySR}: Interpretable Machine Learning for Science with {PySR} and {SymbolicRegression.jl}},
  author       = {Cranmer, Miles},
  year         = {2023},
  eprint       = {2305.01582},
  archivePrefix = {arXiv},
  note   = {arXiv:2305.01582},
  primaryClass = {astro-ph.IM}
}

@misc{jin2020,
  title        = {{Bayesian} Symbolic Regression},
  author       = {Jin, Ying and Fu, Weilin and Kang, Jian and Guo, Jiadong and Guo, Jian},
  year         = {2020},
  eprint       = {1910.08892},
  archivePrefix = {arXiv},
  note   = {arXiv:1910.08892},
  primaryClass = {stat.ME}
}

@article{bomarito2025,
  author  = {Bomarito, Geoffrey and Leser, Patrick},
  title   = {{Bayesian} symbolic regression via posterior sampling},
  journal = {Philosophical Transactions of the Royal Society A: Mathematical, Physical and Engineering Sciences},
  volume  = {384},
  number  = {2317},
  pages   = {20240590},
  year    = {2026}
}

@InProceedings{allamanis2017,
  title     = {Learning Continuous Semantic Representations of Symbolic Expressions},
  author    = {Allamanis, Miltiadis and Chanthirasegaran, Pankajan and Kohli, Pushmeet and Sutton, Charles},
  booktitle = {Proceedings of the 34th International Conference on Machine Learning},
  pages     = {80--88},
  year      = {2017},
  editor    = {Precup, Doina and Teh, Yee Whye},
  volume    = {70},
  series    = {Proceedings of Machine Learning Research},
  publisher = {PMLR}
}

@inproceedings{dascoli2023,
  title     = {{ODEFormer}: Symbolic Regression of Dynamical Systems with Transformers},
  author    = {d'Ascoli, St{\'e}phane and Becker, S{\"o}ren and Schwaller, Philippe and Mathis, Alexander and Kilbertus, Niki},
  booktitle = {The Twelfth International Conference on Learning Representations},
  year      = {2024}
}

@misc{valipour2021,
  title        = {{SymbolicGPT}: A Generative Transformer Model for Symbolic Regression},
  author       = {Valipour, Mojtaba and You, Bowen and Panju, Maysum and Ghodsi, Ali},
  year         = {2021},
  eprint       = {2106.14131},
  archivePrefix = {arXiv},
  note   = {arXiv:2106.14131},
  primaryClass = {cs.LG}
}

@inproceedings{shojaee2023,
  title     = {Transformer-based Planning for Symbolic Regression},
  author    = {Shojaee, Parshin and Meidani, Kazem and Farimani, Amir Barati and Reddy, Chandan K.},
  booktitle = {Thirty-seventh Conference on Neural Information Processing Systems},
  year      = {2023}
}

@inproceedings{Petersen2019,
  title     = {Deep symbolic regression: Recovering mathematical expressions from data via risk-seeking policy gradients},
  author    = {Petersen, Brenden K. and Landajuela Larma, Mikel and Mundhenk, Terrell N. and Santiago, Claudio Prata and Kim, Soo Kyung and Kim, Joanne Taery},
  booktitle = {International Conference on Learning Representations},
  year      = {2021}
}

@inproceedings{Petersen2022,
  author    = {Landajuela, Mikel and Lee, Chak Shing and Yang, Jiachen and Glatt, Ruben and Santiago, Claudio P. and Aravena, Ignacio and Mundhenk, Terrell and Mulcahy, Garrett and Petersen, Brenden K.},
  booktitle = {Advances in Neural Information Processing Systems},
  editor    = {S. Koyejo and S. Mohamed and A. Agarwal and D. Belgrave and K. Cho and A. Oh},
  pages     = {33985--33998},
  publisher = {Curran Associates, Inc.},
  title     = {A Unified Framework for Deep Symbolic Regression},
  volume    = {35},
  year      = {2022}
}

@inproceedings{mundhenk2021,
  author    = {Mundhenk, T. Nathan and Landajuela, Mikel and Glatt, Ruben and Santiago, Claudio P. and Faissol, Daniel M. and Petersen, Brenden K.},
  title     = {Symbolic regression via neural-guided genetic programming population seeding},
  year      = {2021},
  publisher = {Curran Associates Inc.},
  address   = {Red Hook, NY, USA},
  booktitle = {Proceedings of the 35th International Conference on Neural Information Processing Systems},
  articleno = {1908},
  series    = {NeurIPS '21}
}

@misc{crochepierre2022,
  title        = {A Reinforcement Learning Approach to Domain-Knowledge Inclusion Using Grammar Guided Symbolic Regression},
  author       = {Crochepierre, Laure and Boudjeloud-Assala, Lydia and Barbesant, Vincent},
  year         = {2022},
  eprint       = {2202.04367},
  archivePrefix = {arXiv},
  note   = {arXiv:2202.04367},
  primaryClass = {cs.LG}
}

@inproceedings{holt2023dgsr,
  author    = {Holt, Samuel and Qian, Zhaozhi and van der Schaar, Mihaela},
  title     = {Deep Generative Symbolic Regression},
  booktitle = {The Eleventh International Conference on Learning Representations},
  publisher = {OpenReview.net},
  year      = {2023}
}

@InProceedings{germain2015,
  title     = {{MADE}: Masked Autoencoder for Distribution Estimation},
  author    = {Germain, Mathieu and Gregor, Karol and Murray, Iain and Larochelle, Hugo},
  booktitle = {Proceedings of the 32nd International Conference on Machine Learning},
  pages     = {881--889},
  year      = {2015},
  editor    = {Bach, Francis and Blei, David},
  volume    = {37},
  series    = {Proceedings of Machine Learning Research},
  publisher = {PMLR}
}

@article{uria2016nade,
  author  = {Uria, Benigno and C{\^o}t{\'e}, Marc-Alexandre and Gregor, Karol and Murray, Iain and Larochelle, Hugo},
  title   = {Neural Autoregressive Distribution Estimation},
  journal = {Journal of Machine Learning Research},
  year    = {2016},
  volume  = {17},
  number  = {205},
  pages   = {1--37}
}

@inproceedings{papamakarios2017maf,
  author    = {Papamakarios, George and Pavlakou, Theo and Murray, Iain},
  booktitle = {Advances in Neural Information Processing Systems},
  editor    = {I. Guyon and U. Von Luxburg and S. Bengio and H. Wallach and R. Fergus and S. Vishwanathan and R. Garnett},
  publisher = {Curran Associates, Inc.},
  title     = {Masked Autoregressive Flow for Density Estimation},
  volume    = {30},
  year      = {2017}
}

@inproceedings{dinh2017,
  title     = {Density estimation using Real {NVP}},
  author    = {Dinh, Laurent and Sohl-Dickstein, Jascha and Bengio, Samy},
  booktitle = {International Conference on Learning Representations},
  year      = {2017}
}

@inproceedings{tran2019,
  author    = {Tran, Dustin and Vafa, Keyon and Agrawal, Kumar and Dinh, Laurent and Poole, Ben},
  booktitle = {Advances in Neural Information Processing Systems},
  editor    = {H. Wallach and H. Larochelle and A. Beygelzimer and F. d\textquotesingle Alch\'{e}-Buc and E. Fox and R. Garnett},
  publisher = {Curran Associates, Inc.},
  title     = {Discrete Flows: Invertible Generative Models of Discrete Data},
  volume    = {32},
  year      = {2019}
}

@inproceedings{hoogeboom2019,
  author    = {Hoogeboom, Emiel and Peters, Jorn and van den Berg, Rianne and Welling, Max},
  booktitle = {Advances in Neural Information Processing Systems},
  editor    = {H. Wallach and H. Larochelle and A. Beygelzimer and F. d\textquotesingle Alch\'{e}-Buc and E. Fox and R. Garnett},
  publisher = {Curran Associates, Inc.},
  title     = {Integer Discrete Flows and Lossless Compression},
  volume    = {32},
  year      = {2019}
}

@article{papamakarios2021,
  author  = {Papamakarios, George and Nalisnick, Eric and Rezende, Danilo Jimenez and Mohamed, Shakir and Lakshminarayanan, Balaji},
  title   = {Normalizing Flows for Probabilistic Modeling and Inference},
  journal = {Journal of Machine Learning Research},
  year    = {2021},
  volume  = {22},
  number  = {57},
  pages   = {1--64}
}

@article{bondtaylor2022,
  title   = {Deep Generative Modelling: A Comparative Review of {VAEs}, {GANs}, Normalizing Flows, Energy-Based and Autoregressive Models},
  volume  = {44},
  number  = {11},
  journal = {IEEE Transactions on Pattern Analysis and Machine Intelligence},
  author  = {Bond-Taylor, Sam and Leach, Adam and Long, Yang and Willcocks, Chris G.},
  year    = {2022},
  pages   = {7327--7347}
}

@InProceedings{campbell2024dfm,
  title     = {Generative Flows on Discrete State-Spaces: Enabling Multimodal Flows with Applications to Protein Co-Design},
  author    = {Campbell, Andrew and Yim, Jason and Barzilay, Regina and Rainforth, Tom and Jaakkola, Tommi},
  booktitle = {Proceedings of the 41st International Conference on Machine Learning},
  pages     = {5453--5512},
  year      = {2024},
  editor    = {Salakhutdinov, Ruslan and Kolter, Zico and Heller, Katherine and Weller, Adrian and Oliver, Nuria and Scarlett, Jonathan and Berkenkamp, Felix},
  volume    = {235},
  series    = {Proceedings of Machine Learning Research},
  publisher = {PMLR}
}

@inproceedings{chen2023,
  title     = {Analog Bits: Generating Discrete Data using Diffusion Models with Self-Conditioning},
  author    = {Chen, Ting and Zhang, Ruixiang and Hinton, Geoffrey},
  booktitle = {The Eleventh International Conference on Learning Representations},
  year      = {2023}
}

@article{karlis2023integer,
  author  = {Karlis, Dimitris and Mamode Khan, Naushad},
  title   = {Models for Integer Data},
  journal = {Annual Review of Statistics and Its Application},
  year    = {2023},
  volume  = {10},
  number  = {1},
  pages   = {297--323}
}

@article{bingham2019pyro,
  author  = {Bingham, Eli and
             Chen, Jonathan P. and
             Jankowiak, Martin and
             Obermeyer, Fritz and
             Pradhan, Neeraj and
             Karaletsos, Theofanis and
             Singh, Rohit and
             Szerlip, Paul A. and
             Horsfall, Paul and
             Goodman, Noah D.},
  title   = {Pyro: Deep Universal Probabilistic Programming},
  journal = {Journal of Machine Learning Research},
  volume  = {20},
  pages   = {28:1--28:6},
  year    = {2019}
}

@article{KDE1,
  author    = {Rosenblatt, Murray},
  title     = {{Remarks on Some Nonparametric Estimates of a Density Function}},
  volume    = {27},
  journal   = {The Annals of Mathematical Statistics},
  number    = {3},
  publisher = {Institute of Mathematical Statistics},
  pages     = {832--837},
  year      = {1956}
}

@article{KDE2,
  author    = {Parzen, Emanuel},
  title     = {{On Estimation of a Probability Density Function and Mode}},
  volume    = {33},
  journal   = {The Annals of Mathematical Statistics},
  number    = {3},
  publisher = {Institute of Mathematical Statistics},
  pages     = {1065--1076},
  year      = {1962}
}

@article{MoM,
  author    = {Pearson, Karl},
  journal   = {Philosophical Transactions of the Royal Society of London. A},
  pages     = {71--110},
  title     = {Contributions to the Mathematical Theory of Evolution},
  volume    = {185},
  year      = {1894}
}

@article{MLE,
  author  = {Fisher, R. A.},
  title   = {On the mathematical foundations of theoretical statistics},
  journal = {Philosophical Transactions of the Royal Society of London, Series A},
  volume  = {222},
  number  = {594--604},
  pages   = {309--368},
  year    = {1922}
}

@article{Pierson2015-uo,
  title   = {{ZIFA}: Dimensionality reduction for zero-inflated single-cell gene expression analysis},
  author  = {Pierson, Emma and Yau, Christopher},
  journal = {Genome Biology},
  volume  = {16},
  number  = {1},
  pages   = {241},
  year    = {2015}
}

@article{Townes2019-id,
  title   = {Feature selection and dimension reduction for single-cell {RNA-Seq} based on a multinomial model},
  author  = {Townes, F. William and Hicks, Stephanie C. and Aryee, Martin J. and Irizarry, Rafael A.},
  journal = {Genome Biology},
  volume  = {20},
  number  = {1},
  pages   = {295},
  year    = {2019}
}

@article{Risso2018-ob,
  title   = {A general and flexible method for signal extraction from single-cell {RNA-seq} data},
  author  = {Risso, Davide and Perraudeau, Fanny and Gribkova, Svetlana and Dudoit, Sandrine and Vert, Jean-Philippe},
  journal = {Nature Communications},
  volume  = {9},
  number  = {1},
  pages   = {284},
  year    = {2018}
}

@article{Zheng2017-zs,
  title   = {Massively parallel digital transcriptional profiling of single cells},
  author  = {Zheng, Grace X. Y. and Terry, Jessica M. and Belgrader, Phillip and
             Ryvkin, Paul and Bent, Zachary W. and Wilson, Ryan and Ziraldo, Solongo B. and
             Wheeler, Tobias D. and McDermott, Geoff P. and Zhu, Junjie and
             Gregory, Mark T. and Shuga, Joe and Montesclaros, Luz and
             Underwood, Jason G. and Masquelier, Donald A. and Nishimura, Stefanie Y. and
             Schnall-Levin, Michael and Wyatt, Paul W. and Hindson, Christopher M. and
             Bharadwaj, Rajiv and Wong, Alexander and Ness, Kevin D. and
             Beppu, Lan W. and Deeg, H. Joachim and McFarland, Christopher and
             Loeb, Keith R. and Valente, William J. and Ericson, Nolan G. and
             Stevens, Emily A. and Radich, Jerald P. and Mikkelsen, Tarjei S. and
             Hindson, Benjamin J. and Bielas, Jason H.},
  journal = {Nature Communications},
  volume  = {8},
  number  = {1},
  pages   = {14049},
  year    = {2017}
}

@misc{tohme2024isrinvertiblesymbolicregression,
  title        = {{ISR}: Invertible Symbolic Regression},
  author       = {Tohme, Tony and Khojasteh, Mohammad Javad and Sadr, Mohsen and Meyer, Florian and Youcef-Toumi, Kamal},
  year         = {2024},
  eprint       = {2405.06848},
  archivePrefix = {arXiv},
  note   = {arXiv:2405.06848},
  primaryClass = {cs.LG}
}

@misc{tohme2024messyestimationmaximumentropybased,
  title        = {{MESSY} Estimation: Maximum-Entropy based Stochastic and Symbolic densit{Y} Estimation},
  author       = {Tohme, Tony and Sadr, Mohsen and Youcef-Toumi, Kamal and Hadjiconstantinou, Nicolas G.},
  year         = {2024},
  eprint       = {2306.04120},
  archivePrefix = {arXiv},
  note   = {arXiv:2306.04120},
  primaryClass = {cs.LG}
}

@book{Gelman2013-gm,
  title     = {Bayesian Data Analysis},
  author    = {Gelman, Andrew and Carlin, John B. and Stern, Hal S. and Dunson, David B. and Vehtari, Aki and Rubin, Donald B.},
  publisher = {Chapman and Hall/CRC},
  year      = {2013}
}

@article{JMLR:v23:21-0335,
  author  = {Lin, Huiming and Li, Meng},
  title   = {Double Spike {Dirichlet} Priors for Structured Weighting},
  journal = {Journal of Machine Learning Research},
  year    = {2022},
  volume  = {23},
  number  = {248},
  pages   = {1--28}
}

@book{Carroll2017,
  author    = {Carroll, R. J. and Ruppert, D.},
  year      = {2017},
  title     = {Transformation and Weighting in Regression},
  publisher = {Chapman and Hall/CRC}
}
